\documentclass[preprint,12pt,number]{elsarticle}

\biboptions{sort&compress}
\usepackage[a4paper,
            bindingoffset=0.2in,
            left=0.5in,
            right=0.5in,
            top=0.7in,
            bottom=0.7in,
            footskip=.25in]{geometry}

\usepackage{array}
\usepackage{booktabs}
\usepackage{multirow}
\usepackage{siunitx}
\usepackage{verbatim}
\usepackage{enumitem}
\usepackage[hidelinks]{hyperref}
\usepackage{adjustbox}
\usepackage{color,soul}
\usepackage{caption}
\usepackage{subcaption}
\usepackage{array}
\usepackage[dvipsnames]{xcolor}
\usepackage[table]{xcolor}

\usepackage[utf8]{inputenc}
\usepackage{mathtools, nccmath}

\usepackage{amsmath, amsfonts}
\usepackage{amssymb}
\usepackage[mathlines]{lineno}
\usepackage{lineno}
\usepackage{makecell}

\usepackage{textcomp}
\usepackage{graphicx}
\usepackage{algorithm} 
\usepackage{algpseudocode}
\usepackage{algorithmicx}
\usepackage[figurename=Fig.]{caption}

\usepackage{wrapfig}
\DeclareCaptionLabelFormat{custom}{\textbf{#1 #2}}
\DeclareCaptionLabelSeparator{custom}{. }
\DeclareCaptionFormat{custom}{#1#2 #3}
\makeatletter
\def\ps@pprintTitle{%
  \let\@oddhead\@empty%
  \let\@evenhead\@empty%
  \def\@oddfoot{}%
  \let\@evenfoot\@oddfoot}

\begin{document}

\begin{frontmatter}

\title{Efficient and Scalable Physics-Guided Fully Convolutional Spatiotemporal Learning for 3D Microstructure Evolution Prediction}

\author{{Michael Trimboli}}

\author{{Wenxi Liu}}

\author{Xianqi Li\corref{cor1}}

\cortext[cor1]{Corresponding authors}

\affiliation{organization={Department of Mathematics and Systems Engineering},
            addressline={Florida Institute of Technology}, 
            city={Melbourne},
            state={FL},
            postcode={32901}, 
            country={USA}}
            
\begin{abstract}

Accurate prediction of three-dimensional (3D) microstructure evolution remains computationally demanding because high-fidelity phase-field simulations require repeated numerical integration over large volumetric domains and long temporal horizons. This study develops an efficient and scalable physics-guided fully convolutional spatiotemporal framework for direct multi-frame prediction of complete 3D microstructure sequences. The model combines shared 3D spatial encoding and decoding with a factorized latent translator that integrates temporal, local 3D spatial, and channel interactions. A discrete Cahn--Hilliard (CH) residual is incorporated during training to regularize the learned evolution toward the governing dynamics without altering the inference pathway. The framework is evaluated on high-resolution 3D spinodal-decomposition trajectories under nominal, long-horizon, and reduced-temporal-context forecasting. Under full temporal context, the model accurately reproduces volumetric evolution, with average 3D structural similarity remaining above 0.97 over the nominal prediction horizon. Physics guidance becomes increasingly beneficial as temporal information is reduced, improving predictive robustness and preservation of interface-level morphology. The framework also achieves more than a 30-fold wall-clock speedup relative to the reference spectral phase-field solver, while physics guidance introduces no additional inference cost. These results establish direct multi-frame, physics-guided fully convolutional learning as a high-throughput surrogate strategy for dense 3D phase-field dynamics and repeated microstructure forecasting.
\end{abstract}


\end{frontmatter}

\section{Introduction}
\label{sec:introduction}

Microstructure evolution plays a central role in
processing--structure--property relationships in materials. Changes in
grain size, phase distribution, interfacial geometry, connectivity, and
characteristic length scale can strongly affect macroscopic properties
and performance
\cite{chen2002phasefield,rohrer2005interface,
deschamps2021precipitation}. Representative processes include grain
growth, precipitation and coarsening, spinodal decomposition, and
solidification
\cite{voorhees1992ostwald,oono1988spinodal,
boettinger2002solidification}. Accurate prediction of these evolving
structures is therefore important for understanding materials behavior
and for enabling accelerated materials design, process optimization,
uncertainty analysis, and data-integrated materials modeling.

Phase-field methods provide a versatile computational framework for
describing microstructure evolution
\cite{chen2002phasefield,tourret2022phasefield}. By representing
microstructures through continuous order parameters governed by
nonlinear partial differential equations, phase-field models naturally
capture diffuse interfaces, topological transitions, phase separation,
grain-boundary migration, and coarsening without explicit interface
tracking. They have been extensively used to study grain growth
\cite{krill2002grain3d}, diffusion-controlled coarsening
\cite{lifshitz1961kinetics,wagner1961theorie,
voorhees1992ostwald,akaiwa1994late,wang2024systematic}, solidification
\cite{boettinger2002solidification,nestler2000multiphase}, and thin-film
microstructure evolution \cite{stewart2020thinfilm}. For spinodal
decomposition, the Cahn--Hilliard (CH) equation describes the evolution
of a conserved concentration field driven by chemical free-energy
reduction and interfacial-energy effects
\cite{cahn1958freeenergy,cahn1961spinodal}.

The physical fidelity of phase-field simulation, however, comes at
substantial computational cost. High-resolution simulations require
repeated numerical integration of nonlinear PDEs over fine spatial
grids and long temporal horizons. Considerable effort has therefore
been devoted to high-performance computing, parallel implementation,
large-scale GPU simulation, and advanced numerical techniques
\cite{shimokawabe2011dendrite,hunter2011phasefield3d,
vondrous2014parallel,miyoshi2017ultralarge,shi2017gpu,
du2020phasefield}. Although these advances have enabled increasingly
large and realistic simulations, repeated numerical integration remains
expensive in many-query settings involving parameter exploration,
inverse analysis, uncertainty quantification, optimization, and
repeated forward prediction. This challenge becomes particularly severe
for high-resolution 3D systems because the number of spatial degrees
of freedom grows rapidly with grid resolution.

Machine learning surrogate models provide a complementary strategy for
reducing this computational burden. Early studies investigated
learning-based approximations of microstructure evolution and
high-dimensional materials dynamics
\cite{zhang2020multiresolution,montesdeoca2021surrogate}.
Convolutional recurrent models subsequently demonstrated successful
forecasting of grain growth and spinodal decomposition
\cite{yang2021selfsupervised,farizhandi2023spatiotemporal}, while
autoencoder--recurrent formulations learned the temporal evolution of
compressed latent representations
\cite{hu2022latent,ahmad2023autoencoder}. Neural operator methods offer
another route by directly learning mappings between functions
\cite{oommen2022learning,li2020fno}. More broadly, recurrent
spatiotemporal architectures such as PredRNN++ and E3D-LSTM have shown
strong sequence-modeling capability
\cite{wang2018predrnn,wang2018e3dlstm}, while physics-informed neural
networks provide a general framework for incorporating governing
equations into learning \cite{raissi2019pinn}. These developments
demonstrate the feasibility of data-driven surrogate modeling for
microstructure evolution, but challenges remain in computational
efficiency, long-horizon stability, high-resolution prediction, and
physical consistency.

Our previous work introduced a fully convolutional, nonrecurrent
spatiotemporal framework for grain growth and spinodal decomposition
\cite{trimboli2026fully}. Rather than propagating a recurrent hidden
state frame by frame, the model directly learned multi-frame evolution
through an encoder--translator--decoder architecture. The framework
captured both short-term morphology evolution and longer-term
statistical behavior while substantially reducing inference cost
relative to recurrent architectures. Its fully convolutional structure
also enabled direct transfer from lower- to higher-resolution
microstructures without architectural modification or retraining,
demonstrating strong resolution scalability. We subsequently extended
this framework by incorporating the CH equation as a differentiable
physics residual during training
\cite{trimboli2026physicsguided}. The physics-guided formulation
improved robustness under extended forecasting and limited temporal
context while retaining the same efficient inference pathway.
Together, these studies established the complementary benefits of
fully convolutional spatiotemporal forecasting and equation-based
physical regularization.

Both previous studies, however, were restricted to 2D microstructures.
Moving from 2D spatiotemporal prediction to 3D volumetric spatiotemporal prediction is not simply an increase in image dimensionality. A 2D field represents only a planar section of
an evolving material and generally does not uniquely characterize its
volumetric topology. Phase connectivity, interfacial area and
curvature, grain-neighborhood relationships, transport pathways, and
dendritic branching are intrinsically 3D quantities. The importance of
full 3D representation has been demonstrated in phase-field studies of
grain growth \cite{krill2002grain3d,miyoshi2017ultralarge},
experimentally reconstructed dendritic structures
\cite{alkemper2001dendritic}, large-scale dendritic solidification
\cite{shimokawabe2011dendrite}, and quantitative 3D CH simulations of
spinodal decomposition in Fe--Cr alloys
\cite{zhang2024fecr3d}. Complete 3D forecasting therefore provides
structural and topological information that cannot, in general, be
recovered from independently predicted 2D sections.

Learning 3D evolution also introduces substantially greater
computational and modeling challenges. A volumetric field contains far
more degrees of freedom than a 2D field, and temporal evolution must be
learned while preserving through-plane connectivity and volumetric
interface morphology. Recent studies have approached this problem using
several different representations. GrainGNN models 3D grain structures
as dynamic graphs and predicts topological and interfacial evolution in
a reduced graph representation \cite{qin2024graingnn}. Fan et al.\
developed graph neural network surrogates with adaptive spatiotemporal
resolution for 2D and 3D microstructure evolution
\cite{fan2024gnn}. Lanzoni et al.\ introduced a physics-inspired
convolutional recurrent network for 3D CH evolution and demonstrated
long-time extrapolation together with thermodynamic consistency
\cite{lanzoni2024extreme}. More recently, Razavi and Moelans combined
convolutional autoencoding, graph convolution, LSTM-based temporal
modeling, and physics-informed constraints for long-horizon 2D and 3D
microstructure prediction \cite{razavi2026gcnlstm}. These studies
establish the feasibility of learning 3D microstructure dynamics using
graph-based, recurrent, and physics-informed representations. However,
the combination of native-grid volumetric prediction, nonrecurrent
multi-frame forecasting, factorized temporal–3D-spatial–channel learning, and equation-based physical guidance remains comparatively underexplored.

This motivates a representation that separates temporal evolution from
volumetric spatial interactions without resorting to monolithic 4D
operations. Fully convolutional predictive learning provides an
attractive foundation because it avoids recurrent hidden-state
propagation and enables parallel multi-frame prediction. Extending this concept to 3D
microstructure evolution, however, requires a latent representation
that can efficiently capture temporal progression, local volumetric
interactions, and nonlinear feature coupling. 
\textit{Motivated by these considerations}, we develop an efficient and scalable
physics-guided 3D fully convolutional spatiotemporal framework for complete
volumetric microstructure evolution prediction. The proposed framework builds on the fully convolutional encoder--translator--decoder paradigm of SimVP and SimVPv2 \cite{gao2022simvp,tan2025simvpv2}, while reformulating the spatial representation and latent dynamics for native-grid 3D microstructure evolution. In particular, the latent translator factorizes temporal, local 3D spatial, and channel interactions to avoid explicit 4D convolution while maintaining efficient multi-frame volumetric prediction. Temporal operations capture evolution across
neighboring states, 3D convolutions represent local volumetric
interactions and interface morphology, and channel mixing provides
nonlinear coupling among learned features. This factorization preserves
the native 3D organization of the microstructure while controlling the
memory and computational cost of spatiotemporal learning.

Physics guidance is incorporated through the governing 3D CH dynamics.
A differentiable CH residual is evaluated on predicted volumetric
sequences and used to regularize the learned temporal evolution toward
the governing dynamics. Because the physics term is evaluated only
during training, the physics-guided and data-driven variants share the
same inference architecture. Thus, physics guidance modifies the
training objective without introducing additional inference-time
computation. 
The main contributions of this work are summarized as follows:

\begin{itemize}
\item We formulate 3D microstructure forecasting as a finite-horizon
volumetric evolution problem, in which an observed sequence is mapped
directly to a block of consecutive future 3D fields, which enables
nonrecurrent multi-frame prediction of complete native-grid
microstructure trajectories rather than one-step state advancement.

\item We develop a factorized latent spatiotemporal operator that
separates temporal evolution, local 3D spatial interactions, and
channel-wise feature coupling, which provides an efficient
representation of coupled volumetric and temporal dynamics while
preserving the native 3D organization of the microstructure.

\item We incorporate the governing CH dynamics through a
discrete residual evaluated on the predicted volumetric sequence,
providing equation-based regularization of the learned evolution
without modifying the inference architecture or increasing
inference-time cost.

\item We demonstrate high-throughput 3D forecasting across nominal,
extended-horizon, and reduced-temporal-context settings, together with
favorable computational scaling across volumetric resolutions and more
than a 30-fold wall-clock speedup over the reference spectral
phase-field solver.
\end{itemize}
The remainder of the paper is organized as follows.
Section~\ref{sec:PFM} introduces the 3D CH formulation and its role in
physics-guided learning. Section~\ref{sec:method_3d} presents the
proposed fully convolutional spatiotemporal architecture and learning
objective. Section~\ref{sec:exp} describes the simulation data,
implementation, and evaluation metrics. Section~\ref{sec:res}
presents the numerical results, followed by discussion and conclusions
in Sections~\ref{sec:disc} and~\ref{sec:conc}.

\section{3D Cahn--Hilliard Dynamics and Physics Guidance}
\label{sec:PFM}

\subsection{Thermodynamic formulation of the 3D Cahn--Hilliard model}

The microstructure evolution considered in this work is governed by the 3D CH equation, a conserved phase-field model widely used to describe phase separation and subsequent coarsening in binary systems. 

Let \(c(\mathbf{r},t)\) denote the local concentration field over a 3D periodic domain \(\Omega\). The total free-energy functional is written as
\begin{equation}
\mathcal{F}[c]
=
\int_{\Omega}
\left[
f(c)
+
\frac{\kappa}{2}
\lvert\nabla c\rvert^2
\right]
\,\mathrm{d}\mathbf{r},
\label{eq:free_energy}
\end{equation}
where \(f(c)\) is the homogeneous chemical free-energy density and \(\kappa>0\) is the gradient-energy coefficient. The first term favors thermodynamically stable bulk phases, whereas the gradient term penalizes rapid spatial variations in composition and gives rise to diffuse interfaces with finite interfacial energy. 

In this model, the homogeneous contribution is represented by the symmetric double-well potential
\begin{equation}
f(c)
=
Wc^2(1-c)^2,
\label{eq:double_well}
\end{equation}
where \(W>0\) controls the height of the energy barrier between the two equilibrium phases. The two minima at \(c=0\) and \(c=1\) represent the preferred bulk states, while intermediate compositions can become thermodynamically unstable. 

Within the spinodal region, \(f''(c)<0\), so infinitesimal concentration fluctuations reduce the homogeneous free energy and are therefore amplified rather than suppressed. This instability initiates spontaneous phase separation. As the phases become established, the evolution transitions from early-stage fluctuation growth to interface formation and subsequently to coarsening, during which interfacial area is progressively reduced. 
The chemical potential is obtained from the variational derivative of the free-energy functional,
\begin{equation}
\mu
=
\frac{\delta\mathcal{F}}{\delta c}
=
f'(c)-\kappa\nabla^2c,
\label{eq:chemical_potential_general}
\end{equation}
which yields 
$\mu
=
-\kappa\nabla^2c
+
2W\left(c-3c^2+2c^3\right)$.
For constant mobility \(M>0\), the concentration field evolves according to
\begin{equation}
\frac{\partial c}{\partial t}
=
\nabla\cdot
\left(
M\nabla\mu
\right)
=
M\nabla^2\mu, 
\label{eq:CH_general}
\end{equation}
which becomes
\begin{equation}
\frac{\partial c}{\partial t}
=
M\nabla^2
\left[
-\kappa\nabla^2c
+
2W\left(c-3c^2+2c^3\right)
\right].
\label{eq:CH}
\end{equation}

Eq.~\eqref{eq:CH} is a nonlinear fourth-order diffusion equation. The fourth-order term controls interfacial smoothing, while the nonlinear chemical contribution drives phase separation. Their competition produces the complex interconnected morphologies and curvature-driven coarsening characteristic of spinodal decomposition. From the perspective of data-driven forecasting, these dynamics provide a challenging 3D spatial+1D temporal learning problem because the evolution depends simultaneously on local interface geometry, neighboring phase configurations, and long-term collective coarsening.

\subsection{Conservation and free-energy dissipation}

Two properties of the Cahn--Hilliard equation are particularly relevant to microstructure evolution prediction. First, the order parameter is conserved. Integrating Eq.~\eqref{eq:CH_general} over the periodic domain gives
\begin{equation}
\frac{\mathrm{d}}{\mathrm{d}t}
\int_{\Omega}
c(\mathbf{r},t)
\,\mathrm{d}\mathbf{r}
=
M
\int_{\Omega}
\nabla^2\mu
\,\mathrm{d}\mathbf{r}
=
0,
\label{eq:mass_conservation}
\end{equation}
where the final equality follows from the periodic boundary conditions. Therefore, the spatially averaged composition remains constant throughout the evolution. 
Second, the CH dynamics are dissipative with respect to the free-energy functional. Using
\(
\mu=\delta\mathcal{F}/\delta c
\)
and Eq.~\eqref{eq:CH_general},
\begin{align}
\frac{\mathrm{d}\mathcal{F}}{\mathrm{d}t}
&=
\int_{\Omega}
\mu
\frac{\partial c}{\partial t}
\,\mathrm{d}\mathbf{r}
\nonumber
=
M
\int_{\Omega}
\mu\nabla^2\mu
\,\mathrm{d}\mathbf{r}
\nonumber
=
-
M
\int_{\Omega}
\lvert\nabla\mu\rvert^2
\,\mathrm{d}\mathbf{r}
\leq 0.
\label{eq:energy_dissipation}
\end{align}
Thus, the system evolves toward states of lower free energy while preserving the total composition.

These conservation and dissipation properties are closely connected to the morphological evolution of spinodal decomposition. During coarsening, small domains shrink or merge, interfaces become smoother, and the characteristic domain scale increases as the system reduces its total interfacial energy. Consequently, accurate long-horizon prediction requires more than voxel-wise similarity: the predicted fields should also retain physically meaningful interface evolution and coarsening behavior. These properties motivate evaluating the learned trajectories using morphology- and physics-sensitive measures in addition to voxel-wise reconstruction metrics. In this study, interface-curvature statistics are used as one such morphology-sensitive descriptor.

\subsection{Cahn--Hilliard physics in the learning framework}

The CH model serves two complementary roles in the proposed framework. First, it defines the physical dynamics used to generate the reference 3D trajectories. Second, it provides a governing-equation constraint that can be evaluated directly on predicted concentration fields. For a predicted field \(\widehat c\), the continuous CH residual is
\begin{equation}
\mathcal{R}_{\mathrm{CH}}(\widehat c)
=
\frac{\partial \widehat c}{\partial t}
-
M\nabla^2
\left[
-\kappa\nabla^2\widehat c
+
2W
\left(
\widehat c-3\widehat c^2+2\widehat c^3
\right)
\right].
\label{eq:CH_continuous_residual}
\end{equation}
A field satisfying the governing equation exactly has
\(\mathcal{R}_{\mathrm{CH}}=0\). In the proposed learning framework, this residual is discretized over the predicted volumetric sequence and incorporated as a physics-based regularizer. 
The physics residual complements the data-fidelity objective rather than replacing it. The data term encourages agreement with the reference trajectory, whereas the CH residual constrains the temporal transition between neighboring predicted states. Because the neural network model learns a spatiotemporal mapping rather than explicitly reproducing the numerical ETD update, visually plausible predictions can gradually depart from the underlying dynamics as forecast errors accumulate. Physics guidance therefore provides an additional dynamical constraint that can be particularly useful for extended forecasting or when temporal observations are limited.

The residual should not be interpreted as exact enforcement of the governing equation. Minimizing \(\mathcal{R}_{\mathrm{CH}}\) encourages consistency with CH dynamics but does not by itself guarantee exact mass conservation or monotonic free-energy dissipation in every predicted sequence. Accordingly, the predictions are evaluated using both conventional reconstruction metrics and morphology-sensitive measures. The discrete formulation and its incorporation into the training objective are described in Section~\ref{sec:method_3d}.

\section{Physics-Guided 3D Fully Convolutional Spatiotemporal Learning Framework}
\label{sec:method_3d}

\subsection{Problem Formulation}
\label{sec:prob_form}

We consider the prediction of full-field 3D microstructure evolution given a finite sequence of volumetric phase-field states as input. Let
$\mathbf{X}_{1:T}
=
\{\mathbf{X}_1,\ldots,\mathbf{X}_{T}\}
\in
\mathbb{R}^{B\times T\times C\times D\times H\times W}$
denote a batch of observed sequences, where $B$ is the batch size, $T$ is the number of input frames, $C$ is the number of physical fields, and $D\times H\times W$ is the volumetric grid. Exact dimensional values are described in Sec.~\ref{sec:DataPre} The target sequence is defined as 
$\mathbf{Y}_{1:T}
=
\mathbf{X}_{T+1:2T}$.
The learned spatiotemporal operator
$\mathcal{F}_{\theta}$ maps an input sequence of length $T$ to a predicted sequence of the same length:
\begin{equation}
\widehat{\mathbf{Y}}_{1:T}
=
\mathcal{F}_{\theta}
\left(
\mathbf{X}_{1:T}
\right).
\label{eq:3d_mapping}
\end{equation}
where $\theta$ represents the estimated parameters. Each time step is a complete 3D field rather than an independently processed 2D slice. The model must therefore preserve through-plane correlations, volumetric connectivity, interface curvature, and coarsening behavior while learning their temporal evolution. This motivates a fully 3D architecture in which all spatial feature extraction and reconstruction are performed with 3D operators. 

\subsection{Architecture Overview}

\begin{figure}[h!]
\centering
\includegraphics[width=\linewidth]{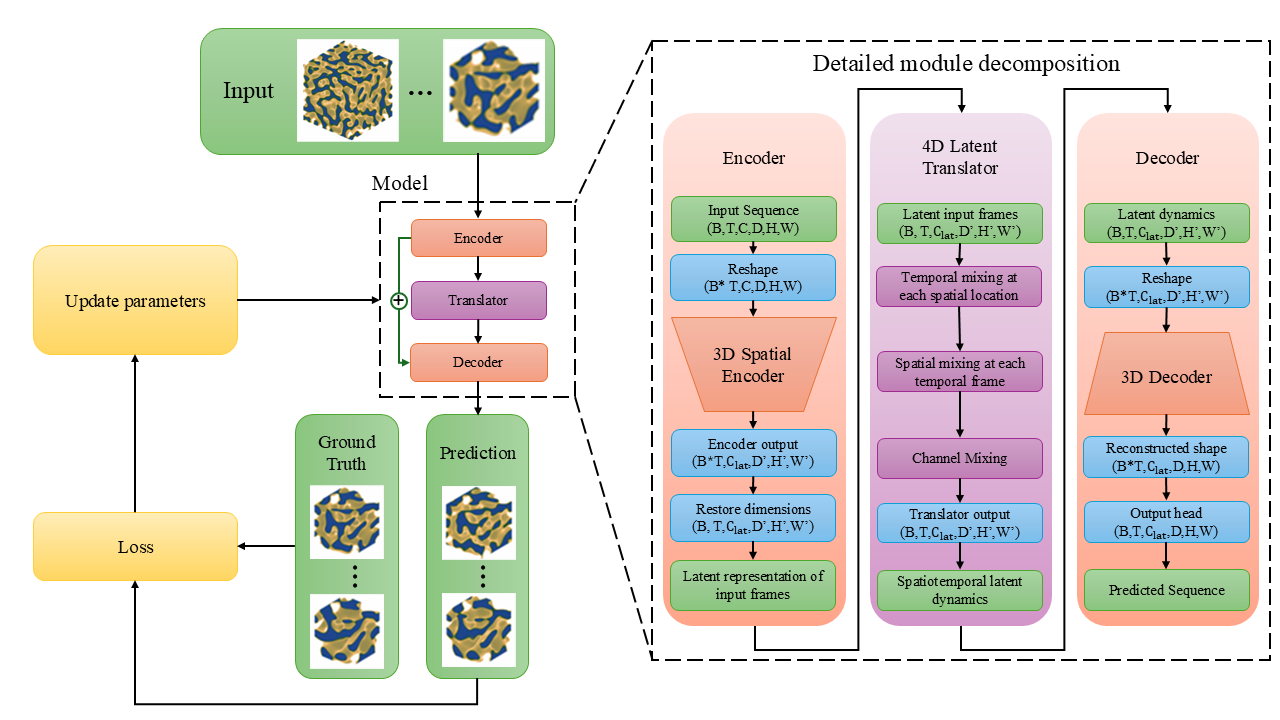}
\caption{{\textbf{Flowchart of proposed architecture:} A 3D microstructure sequence of temporal length $T$ serves as input to the model. The 3D convolutional spatial encoder extracts spatial correlations and phase distributions, producing downsampled latent representations. These latent features are propagated through the 4D latent translator,  encoding temporal evolution patterns and phase separation dynamics. The convolutional spatial decoder then reconstructs the forecast microstructure fields at future timesteps $T+1$ through $2T$. Before the final decoder layer, features from the first encoder layer are added as a residual to the predicted field, represented as a green arrow. To further enhance training, a physics-guided regularization term is incorporated to encourage phase-field-consistent spinodal decomposition dynamics, penalizing deviations from the governing properties.}}
\label{figflow}
\end{figure}

The proposed framework follows a nonrecurrent encoder--translator--decoder structure:
\begin{equation}
\mathcal{F}_{\theta}
=
\mathcal{D}_{\theta_3}
\circ
\mathcal{G}_{\theta_2}
\circ
\mathcal{E}_{\theta_1},
\label{eq:architecture}
\end{equation}
where \(\mathcal{E}_{\theta_1}\) is a shared 3D encoder, \(\mathcal{G}_{\theta_2}\) is a latent spatiotemporal translator, \(\mathcal{D}_{\theta_3}\) is a shared 3D decoder, and $\theta_1, \theta_2, \theta_3$ are parameters in these three components respectively. The tensor flow is
\begin{equation}
\mathbf{X}_{1:T}
\rightarrow
\mathbf{Z}_{1:T}
\rightarrow
\widetilde{\mathbf{Z}}_{1:T}
\rightarrow
\widehat{\mathbf{Y}}_{1:T},
\end{equation}
where 
$\mathbf{Z}_{1:T}, \widetilde{\mathbf{Z}}_{1:T}
\in
\mathbb{R}^{B\times T\times C_h\times D_h\times H_h\times W_h}$.
The encoder extracts volumetric morphology features, the translator models their temporal evolution in a compact latent space, and the decoder reconstructs all future 3D fields. The proposed architecture follows the fully convolutional encoder--translator--decoder design of SimVP/SimVPv2, but is reformulated for volumetric microstructure prediction through shared 3D spatial encoding and decoding and a factorized latent spatiotemporal translator. With this architecture we introduce a modified latent translator and extend sequence forecasting to 3D data. 

Unlike recurrent architectures, the proposed framework does not propagate a hidden state sequentially within each prediction block. The model also operates directly on structured volumetric feature tensors rather than converting the 3D fields into token sequences. This preserves an explicit correspondence between latent spatial locations and the underlying volumetric grid while avoiding the additional representation and computational costs associated with global token interactions. The use of localized convolutional operators further provides a natural inductive bias for microstructure evolution, where interfacial motion and phase interactions are strongly influenced by neighboring spatial regions. At the same time, stacking multiple spatial, temporal, and channel-mixing blocks progressively enlarges the effective receptive field, allowing the network to represent interactions beyond an individual local neighborhood. This design therefore balances computational efficiency with the ability to capture multiscale spatiotemporal dependencies in evolving 3D microstructures.

\subsection{Shared 3D Spatial Encoder}

The encoder is applied independently to each input frame with shared parameters. The batch and temporal dimensions are first merged: 
$\mathbf{X}^{\mathrm{enc}}
=
\operatorname{reshape}
\left(
\mathbf{X}_{1:T}
\right)
\in
\mathbb{R}^{(BT)\times C\times D\times H\times W}$.
This reshaping collapses the batch and temporal dimensions, allowing spatial feature extraction through 2D convolutions. Here, we extend this formulation to volumetric microstructure data by replacing 2D operations with 3D convolutions while preserving temporal independence across frames.

Let \(\mathbf{H}^{(0)}=\mathbf{X}^{\mathrm{enc}}\). The \(i\)-th encoder stage is 
$\mathbf{H}^{(i)}
=
\mathcal{E}^{(i)}
\left(
\mathbf{H}^{(i-1)}
\right),
\quad
i=1,\ldots,N_e$, 
with
\begin{equation}
\mathcal{E}^{(i)}(\mathbf{H})
=
\sigma
\left[
\mathcal{N}
\left(
\operatorname{Conv3D}^{(i)}_2
\left(
\sigma
\left[
\mathcal{N}
\left(
\operatorname{Conv3D}^{(i)}_1(\mathbf{H})
\right)
\right]
\right)
\right)
\right].
\label{eq:encoder_stage}
\end{equation}
Batch normalization and GELU activation are used within spatial convolutional stages to stabilize feature distributions during training. Strided 3D convolutions are introduced at selected stages to reduce the spatial dimensions. After \(m\) downsampling stages,
$D_h\approx\frac{D}{2^m},
\quad
H_h\approx\frac{H}{2^m},
\quad
W_h\approx\frac{W}{2^m}$. 
The encoder output is reshaped to
\begin{equation}
\mathbf{Z}_{1:T}
\in
\mathbb{R}^{B\times T\times C_h\times D_h\times H_h\times W_h}.
\end{equation}
Merging the batch and temporal dimensions allows the encoder to extract 3D spatial features from all input frames in parallel using shared convolutional operations, while temporal evolution is subsequently modeled by the latent spatiotemporal translator. The encoder therefore serves two primary functions: it captures local 3D morphology, including diffuse interfaces, curved boundaries, and volumetric domain connectivity, and it compresses the spatial representation to a lower-resolution latent space. This spatial compression substantially reduces the memory and computational cost of subsequent spatiotemporal modeling, which is particularly important for high-resolution 3D microstructure data.

\subsection{Latent 4D Spatiotemporal Translator}

The proposed architecture of the translator mixes microstructural information across \(3D+1D\) to capture dynamical evolution processes. The translator consists of three components: temporal mixing kernels,  spatial mixing kernels, and channel mixing kernels. The translator consists of \(N_g\) residual spatiotemporal mixing blocks:
\begin{equation}
\begin{aligned}
\mathbf{Z}^{(l)}_{T} &= \mathbf{Z}^{(l)} + \mathcal{T}^{(l)}\!\left(\mathrm{GN}_t(\mathbf{Z}^{(l)})\right), \\
\mathbf{Z}^{(l)}_{TS} &= \mathbf{Z}^{(l)}_{T} + \mathcal{S}^{(l)}\!\left(\mathrm{GN}_s(\mathbf{Z}^{(l)}_{T})\right), \\
\mathbf{Z}^{(l+1)} &= \mathbf{Z}^{(l)}_{TS} + \mathcal{C}^{(l)}\!\left(\mathrm{GN}_c(\mathbf{Z}^{(l)}_{TS})\right).
\end{aligned}
\label{eq:translator_block}
\end{equation}

Here, \(\mathcal{T}^{(l)}\) learns the temporal modulation of the latent features, \(\mathcal{S}^{(l)}\) learns local three-dimensional interactions, and \(\mathcal{C}^{(l)}\) models interactions between spatiotemporal features. Group normalization is applied before each operation to stabilize features without dampening instance-specific structures.

Spatiotemporal relations are captured in the translator for the 3D decoder to extract. Temporal mixing models the evolution of these latent features across time, effectively learning a discrete-time approximation of the underlying dynamical system. Spatial mixing, in contrast, captures local spatial interactions within each latent state along the volumetric grid. Channel mixing learns how these latent features combine at every spatiotemporal point. Together, these operators factorize spatiotemporal modeling into complementary components. This decomposition enables efficient learning of \(3D+1D\) dynamical evolution without explicit 4D convolution or global attention. To represent successive application of temporal, spatial, and channel mixing operators, superscripts $(T)$, $(TS)$, and $(TSC)$ are used respectively.

\subsection{Lightweight 4D Temporal Mixing}

Temporal mixing is performed using a depthwise separable one-dimensional convolution along the temporal axis. Given a latent tensor 
$\mathbf{Z}_{1:T}
\in
\mathbb{R}^{B\times T\times C_h\times D_h\times H_h\times W_h}$, 
the operation is applied independently at each spatial location $(D_h,H_h,W_h)$.

We first reshape the tensor such that each spatial position corresponds to an independent temporal sequence:
\begin{equation}
\mathbf{Z}_{b,:,:,d,h,w} \in \mathbb{R}^{T \times C_h}.
\end{equation}
A depthwise 1D temporal convolution captures local temporal neighborhoods independently within each latent channel: 
$\hat{\mathbf{Z}}_{b,t,c,d,h,w} = \sum_{\tau \in \mathcal{N}(t)} 
K^{(t)}_c(\tau) \, \mathbf{Z}_{b,t-\tau,c,d,h,w}$, 
where $K^{(t)}_c$ denotes a learnable temporal kernel for channel $C_h$.

This is followed by a pointwise convolution to enable channel interaction:
$\tilde{\mathbf{Z}}_{b,t,:,d,h,w} = W^{(t)} \hat{\mathbf{Z}}_{b,t,:,d,h,w}$. 

A nonlinearity is then applied:
$\mathbf{Z}_{1:T}^{(t)} = \mathrm{GELU}(\tilde{\mathbf{Z}})$. 

The superscript $(t)$ denotes the latent representation after temporal mixing. The temporal convolution aggregates information from neighboring time steps, enabling multi-frame feature extraction at each spatial location. This design allows the translator to model temporal correlations and infer the evolution of the material.

\subsection{Efficient 4D Spatial Mixing}

Spatial mixing is implemented using a 3D convolution applied independently at each time step. Given 
$\mathbf{Z}_{1:T}^{(t)}
\in
\mathbb{R}^{B\times T\times C_h\times D_h\times H_h\times W_h}$, 
we process each temporal slice separately. The spatial operator is defined as:
\begin{equation}
\mathbf{Z}_{b,t,:,:,:,:}^{(TS)} = 
\phi\big(
\mathrm{BN}\big(
\mathrm{Conv3D}(\mathbf{Z}_{b,t,:,:,:,:}^{(T)})
\big)\big),
\end{equation}
where $\mathrm{Conv3D}$ uses a kernel of size $k_s \times k_s \times k_s$, $\mathrm{BN}$ denotes batch normalization, and $\phi$ is the GELU activation. The superscript $(S)$ denotes the latent representation passed through the spatial mixing kernels. 

This operation captures local spatial correlations within each time step, enabling the model to learn interactions across neighboring voxels in the $(D,H,W)$ domain while preserving temporal structure.

\subsection{Pointwise Channel Mixing (4D)}

Channel mixing is performed using a pointwise MLP applied independently at each spatiotemporal location. Analogous to the channel-mixing MLP in MLP-Mixer \cite{tolstikhin2021mlp}, this is implemented using $1 \times 1 \times 1$ convolutions defined as:

\begin{equation}
\mathbf{Z}^{(TSC)}_{1:T} =
W_2^{(c)} \, \mathrm{GELU}\left(
W_1^{(c)}  \mathbf{Z}_{1:T}^{(TS)}
\right),
\end{equation}
where $W_1^{(c)} \in \mathbb{R}^{C' \times C_h}$, $W_2^{(c)} \in \mathbb{R}^{C_h \times C'}$ are learnable parameters, and $C' = \alpha C_h$ is an expanded channel dimension by factor $\alpha.$ The superscript $(C)$ denotes the latent representation passed through the MLP mixing kernels. This module applies position-wise channel mixing across the latent tensor, enabling nonlinear interactions between feature channels at each spatial location and timestep, thereby enriching the local spatiotemporal representation without performing temporal projection or spatial resampling, and without disrupting the underlying topology.

The spatial branch therefore preserves local 3D morphology, the  temporal branch captures temporal dependencies, and the channel branch enhances interactions between the aforementioned extracted features. Their repeated residual interaction models coupled \(3D+1D\) dynamics without the cost of full 4D convolution or global spatiotemporal attention.

\subsection{Shared 3D Spatial Decoder}

The translator output 
$\widetilde{\mathbf{Z}}_{1:T}
=
\mathbf{Z}^{(TSC)}_{1:T}
=
\widetilde{\mathbf{Z}}^{(N_g)}$ 
is reshaped to 
$\widehat{\mathbf{Z}}^{\mathrm{dec}}
\in
\mathbb{R}^{(BT_{\mathrm{in}})\times C_h\times D_h\times H_h\times W_h}$.
The decoder uses shared 3D upsampling stages:
\begin{equation}
\mathbf{Q}^{(j)}
=
\sigma
\left[
\mathcal{N}
\left(
\operatorname{Conv3D}^{(j)}
\left(
\operatorname{Up3D}^{(j)}
\left(
\mathbf{Q}^{(j-1)}
\right)
\right)
\right)
\right],
\qquad
j=1,\ldots,N_d.
\label{eq:decoder_stage}
\end{equation}
where Up3D is trilinear interpolation. A residual skip connection from the first encoder block is incorporated before the final decoder stage to recover high-resolution spatial information lost during downsampling. After restoring the original spatial resolution, a \(1\times1\times1\) convolution maps the hidden channels to the physical field: 
$\widehat{\mathbf{Y}}^{\mathrm{dec}}
=
\operatorname{Conv3D}_{1\times1\times1}
\left(
\mathbf{Q}^{(N_d)}
\right)$. 
Finally, we have
\begin{equation}
\widehat{\mathbf{Y}}_{1:T}
\in
\mathbb{R}^{B\times T\times C\times D\times H\times W}.
\end{equation}
\subsection{Physics-Guided Training Objective}

The data-fidelity term is the mean squared error over all predicted voxels and time steps:
\begin{equation}
\mathcal{L}_{\mathrm{data}}
=
\frac{1}{BTDHW}
\sum_{b=1}^{B}
\sum_{t=1}^{T}
\left\|
\widehat{\mathbf{Y}}_{b,t}
-
\mathbf{Y}_{b,t}
\right\|_2^2.
\label{eq:data_loss}
\end{equation}
The 3D reference trajectories are generated from the CH equation with a quartic double-well free energy and constant mobility. For each predicted phase field \(\widehat{c}_{b,t}\), the discrete chemical potential is
\begin{equation}
\widehat{\mu}_{b,t}
=
-\kappa\Delta_h\widehat{c}_{b,t}
+
2W
\left(
\widehat{c}_{b,t}
-
3\widehat{c}_{b,t}^{\,2}
+
2\widehat{c}_{b,t}^{\,3}
\right).
\label{eq:chemical_potential}
\end{equation}
The discrete CH residual is
\begin{equation}
\mathcal{R}_{b,t}
=
\frac{
\widehat{c}_{b,t+1}
-
\widehat{c}_{b,t}
}{\Delta t}
-
M\Delta_h\widehat{\mu}_{b,t},
\qquad
t=1,\ldots,T-1.
\label{eq:ch_residual}
\end{equation}
The physics loss is
\begin{equation}
\mathcal{L}_{\mathrm{CH}}
=
\frac{1}{B(T-1)DHW}
\sum_{b=1}^{B}
\sum_{t=1}^{T-1}
\left\|
\mathcal{R}_{b,t}
\right\|_2^2.
\label{eq:physics_loss}
\end{equation}
The total objective is
\begin{equation}
\mathcal{L}
=
\mathcal{L}_{\mathrm{data}}
+
\lambda_{\mathrm{phy}}
\mathcal{L}_{\mathrm{CH}},
\label{eq:total_loss}
\end{equation}
where \(\lambda_{\mathrm{phy}}\) controls the strength of physics regularization. The value for \(\Delta t\) is 1 in Eq.~\ref{eq:ch_residual}, which represents the normalized temporal spacing between consecutive prediction frames. The physics term is therefore formulated as a relative Cahn--Hilliard consistency constraint in normalized frame time, rather than as an exact reproduction of the \texttt{SpectralETD} trajectory. This normalization separates the temporal discretization used by the neural network from the numerical integration timestep of the underlying Cahn--Hilliard solver. Since consecutive saved frames are separated by 10 simulation-time units (see Sec.~\ref{sec:DataGen}), the normalized time coordinate is related to the simulation time by \(t=t_{\mathrm{sim}}/10\). The data-driven baseline is obtained by setting \(\lambda_{\mathrm{phy}}=0\).

The spatial operator \(\Delta_h\) uses periodic boundary treatment consistent with the Fourier pseudo-spectral CH simulations. The physics residual is evaluated only during training; therefore, physics guidance introduces no additional inference-time operations. The exact use of this physics loss term during training is defined in Sec.~\ref{netArch}.

\section{Experimental Setting}
\label{sec:exp}

\subsection{Data Generation}
\label{sec:DataGen}

All microstructure trajectories are generated using \texttt{SpectralETD} version 0.0.4 \cite{soares2023exponentialintegratorsphasefieldequations}. 
The governing parameters are \(W = 1.0, M = 1.0,\) and \(\kappa = 0.01.\) A total of 100 trajectories are generated, each on a grid of size \(128 \times 128 \times 128\). The solver timestep is \(\Delta{t} = 0.01,\) and frames are saved every 10 time units. Consequently, 201 temporal frames are recorded, including the initial condition, corresponding to 2000 units of simulated time. There are a total of 200,000 solver timesteps in this data simulation. The phase-field values are bounded between 0 and 1. The initial condition is set to 0.5 with additive Gaussian noise applied at each voxel. Hence, all generated trajectories preserve the same governing parameters. The training dataset does not account for variations in the governing paramters. Also, this experimental setting is limited to a single concentration field in a binary setting, not account for multicomponent miscrostructure systems. The dimensions of every generated trajectory is $(T,D,H,W)=(201,128,128,128)$.

\subsection{Dataset Construction and Preprocessing}
\label{sec:DataPre}

From the 100 available microstructure trajectories, 80 are used for training, 10 for validation, and the remaining 10 for testing. All spatiotemporal microstructure sequences are organized as tensors of shape 
$
\mathbb{R}^{B \times T \times C \times D \times H \times W},
$
where $B$ denotes the batch size, $T$ is the temporal length, $C$ is the number of channels, $D,$  $H,$ and $W$ are the spatial dimensions. As outlined in Sec.~\ref{sec:DataGen}, the dimensions of each trajectory is $(T,D,H,W)=(201,128,128,128)$. In this work, all microstructure fields are represented as single-channel grayscale images, such that $C=1$. Each sample is treated as an individual batch with $B=1$. The spatial dimension of every sample is $(D,H,W)=(128,128,128)$, representing the volumetric microstructure.

During training and evaluation, each trajectory is partitioned into overlapping samples using a sliding window scheme. The number of samples extracted from a sequence is given by:
$
\left\lfloor \frac{T_{\mathrm{seq}}-W}{S} \right\rfloor + 1.
$
where \(W\) is the total window length (input and output combined) and \(S\) is the stride between consecutive windows. For the proposed model, the network is trained to predict 10 future volumetric frames from $T=10$ input frames, yielding a total window size of \(W = 20\). A stride of \(S = 5\) is selected. Consequently, each trajectory produces 37 samples, each spanning 20 temporal frames.

In total, 2,960 training samples are generated from the 80 training trajectories, and 370 samples from the 10 validation trajectories. This model is used for all four experiments presented in Section \ref{sec:res}. All experiments use the same 10 data trajectories; the primary difference lies in how each testing dataset is constructed. In all experiments, except for the case in which 40 frames are forecast from 10 input frames (see Section \ref{subsec:10-40}), a total of 370 samples are collected and evaluated. For the excluded experiment, 270 samples are evaluated.

In experiments with reduced temporal input, the following preprocessing scheme is applied. The model expects 10 input frames of a 3D microstructure. Let the number of available input frames be $t < 10.$ To match the required input dimension \((T = 10)\), the \(t\) observed frames are first collected, and zero-padding is prepended to fill the remaining \(10-t\) entries. Thus, while the model receives a tensor of length 10, only \(t\) frames contain physical data.

\subsection{Network Architecture and Training}
\label{netArch}

In the proposed framework, four encoder blocks progressively downsample the spatial resolution from 128$^{3}$ to 32$^{3}$. The encoder alternates between stride-1 and stride-2 3D convolutional blocks, resulting in two spatial downsampling operations: $(128^{3} \rightarrow 64^{3} \rightarrow 32^{3})$. Concurrently, the channel dimension increases from 1 to 16, enabling the extraction and storage of higher-level spatial features. 
Both spatial and temporal kernel sizes are 3. Encoder features a stride of 2 whereas decoder features a stride of 1. There are 4 encoder layers and 4 decoder layers in the model. After downsampling through the encoder, the latent dimensions are (B,T,16,32,32,32).

The model is trained for up to 200 epochs with a batch size of 1. Training is conducted on the $128\times128\times128$ spinodal decomposition dataset using the AI Panther system at the Florida Institute of Technology, equipped with NVIDIA A100 SXM4 GPUs. In addition to a baseline fully convolutional model without physics guidance, a physics-guided variant is trained in which the governing CH dynamics are incorporated into the learning objective.

A learning rate of \(1e-3\) was used, with the physics regularization weighted by \(\lambda_{phy}=1e-8\). The model was optimized using the Adam optimizer with no weight decay, minimizing the sum of the data-driven loss and the weighted physics loss. In this work, the physics loss is evaluated only on the final five predicted frames. Physics regularization is only introduced into the training regime.

\subsection{Performance Metrics}

Prediction accuracy is first assessed using standard image-based metrics, namely root-mean-square error (RMSE), structural similarity index measure (SSIM) \cite{ZWang04}, and peak signal-to-noise ratio (PSNR). These metrics are evaluated at every frame over the prediction horizon.

To further evaluate model performance beyond pixel-level agreement, the interface-segment curvature distribution is compared between ground-truth and predicted trajectories. Because the proposed model is trained without explicit physical guidance, it is important to assess not only whether it produces visually accurate predictions, but also whether it correctly captures spinodal decomposition coarsening dynamics and interfacial evolution in a physically consistent manner. To enable a more rigorous comparison between the ground-truth and predicted interface-segment curvature distributions, the total variation distance \cite{gibbs2002onchoosing} is computed between the two distributions.

For a countable state space $\Omega$, the total variational distance is defined as
\begin{equation}
d_{TV} = \frac{1}{2} \sum_{x \in \Omega} \left| \mu(x) - \nu(x) \right|.
\label{dTV}
\end{equation}
where $\mu(x)$ and $\nu(x)$ are the predicted and ground-truth distributions respectively. Total variation distance is computed between discretized curvature distributions after binning over a shared finite partition of the curvature domain.

\section{Numerical Results}
\label{sec:res}

This section explores four different experimental settings for the proposed framework. For the first experiment, the performance of the original model is evaluated on the nominal task of predicting 10-frame sequences given 10 input frames. In the next three experiments, the performance between the proposed framework and the physics-guided version are compared. The purpose of this comparison is to evaluate how significantly the model results change when physics guidance is introduced into the training framework.
In all experiments with a prediction horizon longer than 10 frames, an iterative rollout scheme is applied to extend the prediction horizon. At each rollout step, the model's previous prediction is used as the input to the network, allowing subsequent future frames to be recursively predicted from the model's own outputs.

\begin{equation}
\widehat{\mathbf{Y}}_{(k-1)T+1:kT}
=
\mathcal{F}_{\theta}
\left(
\widehat{\mathbf{Y}}_{(k-2)T+1:(k-1)T}
\right),
\qquad
\qquad k = 2, 3, ..., K
\label{eq:iterative_rollout}
\end{equation}

To compare the results of this iterative prediction,The predictions generated at each rollout step are temporally concatenated to form a single prediction sequence,

\begin{equation}\widehat{\mathbf{Y}}_{1:KT}
=
\Vert^{K}_{k=1}\widehat{\mathbf{Y}}_{(k-1)T+1:kT},
\end{equation}

in which $\Vert$ denotes temporal concatenation. The resulting prediction sequence is then directly compared with the corresponding ground-truth sequence of equal temporal length.

\begin{figure}[h!]
\centering
\includegraphics[width=0.45\linewidth]{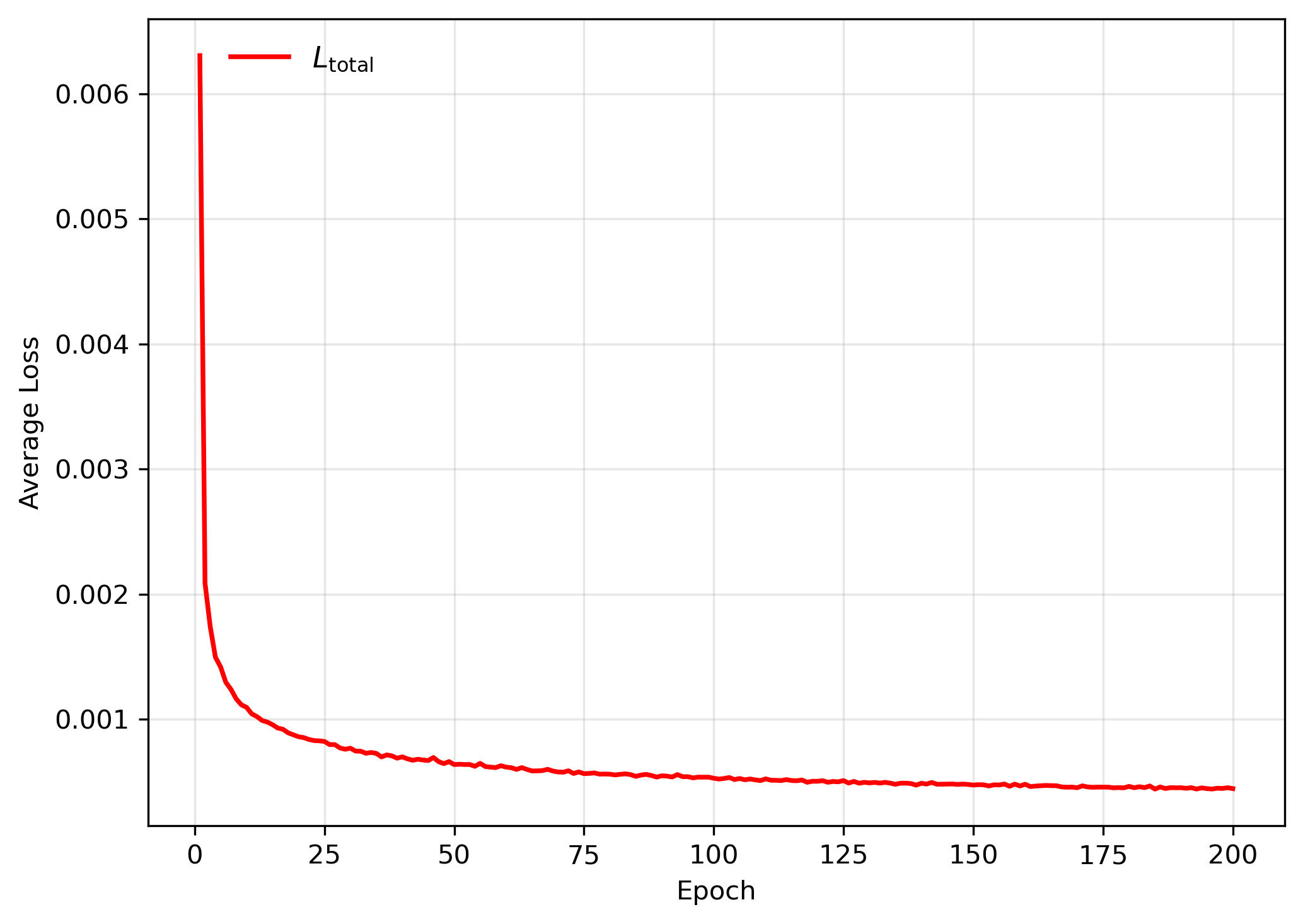}
\caption{\textbf{Physics-guided model total training losses:} Loss contribution to model training is reported at every epoch on the data-driven model.}
\label{epochLoss}
\end{figure}
 
\subsection{Predictions of 10 future frames from 10 input frames (10$\rightarrow$10)}

\begin{figure}[h!]
\centering
\includegraphics[width=0.45\linewidth]{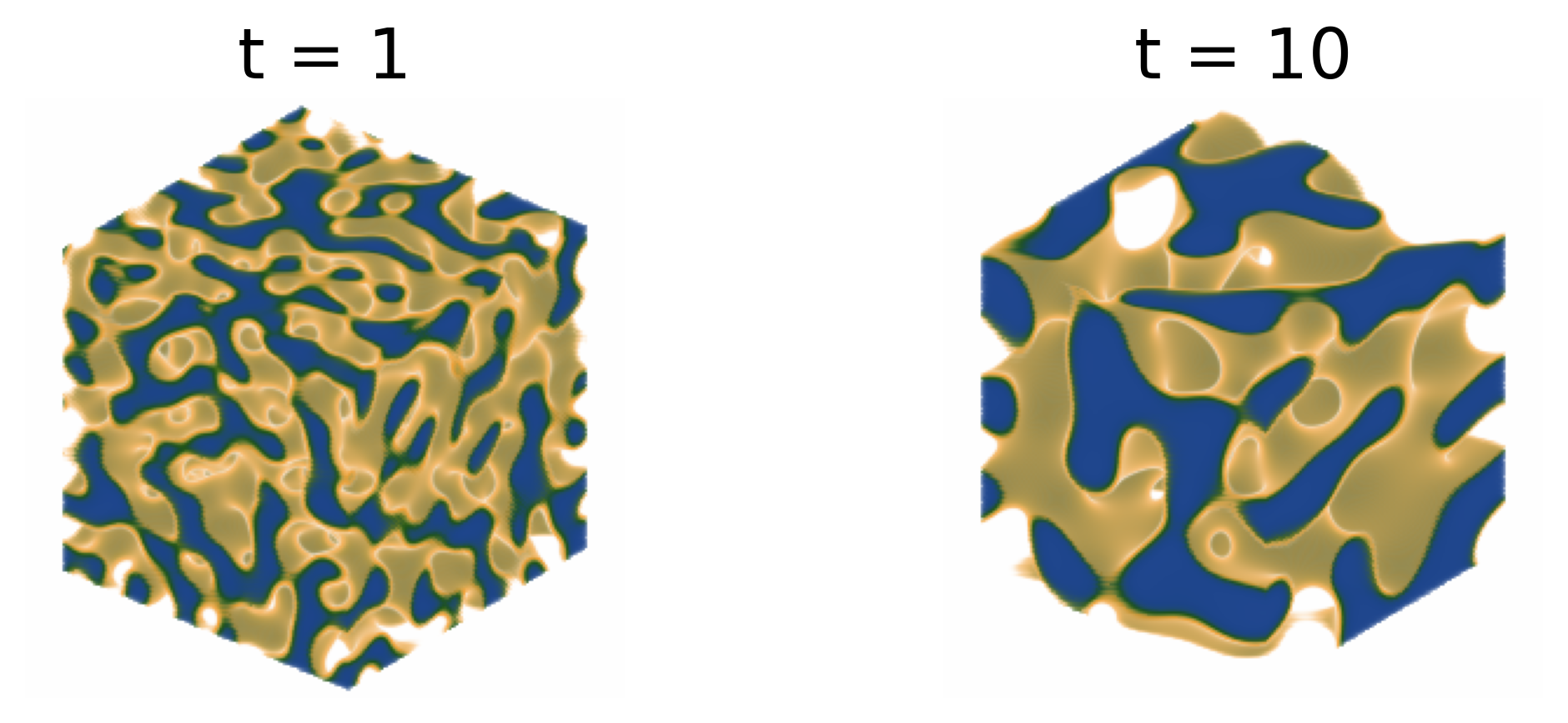}
\caption{\textbf{Spinodal decomposition input (10 input frames, 10 output frames):} Input sequence corresponding to the results shown in Fig.~\ref{fig5}, displayed at timesteps $t = 1, 10$.}
\label{fig4}
\end{figure}

\begin{figure}[h!]
\centering
\includegraphics[width=\linewidth]{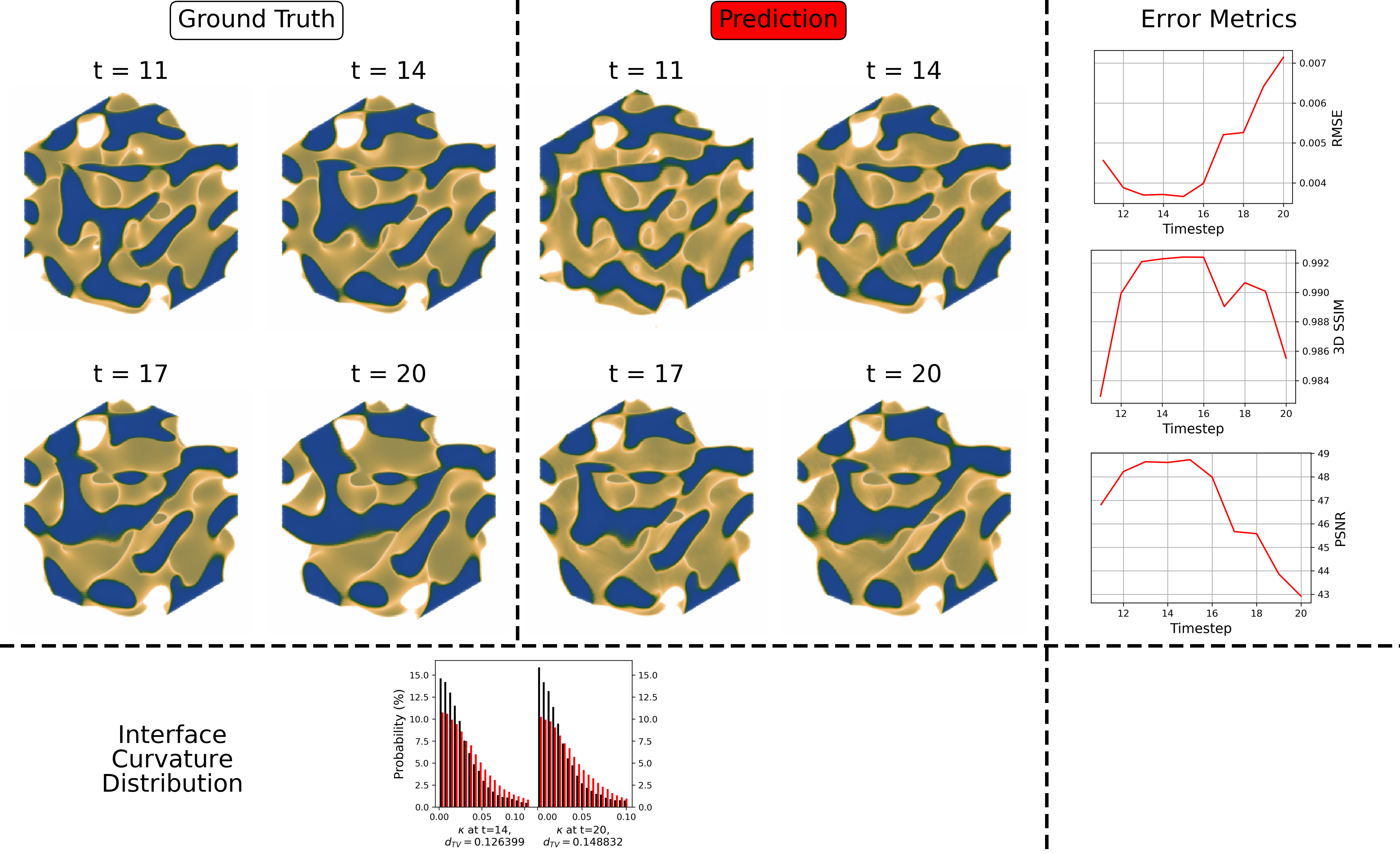}
\caption{\textbf{Spinodal decomposition prediction (10 input frames, 10 output frames) comparison:} Predictions from the proposed model are shown alongside the ground-truth at timesteps $t = 11, 14, 17, 20$. The right column reports the RMSE, 3D SSIM, and PSNR computed between the predicted and ground-truth sequences at each timestep. The bottom row displays the interface curvature distributions for both the prediction and ground-truth at $t = 14, 20$, with the corresponding total variation distance between the distributions provided beneath each distribution.}
\label{fig5}
\end{figure}

\begin{figure}[h!]
\centering
\includegraphics[width=\linewidth]{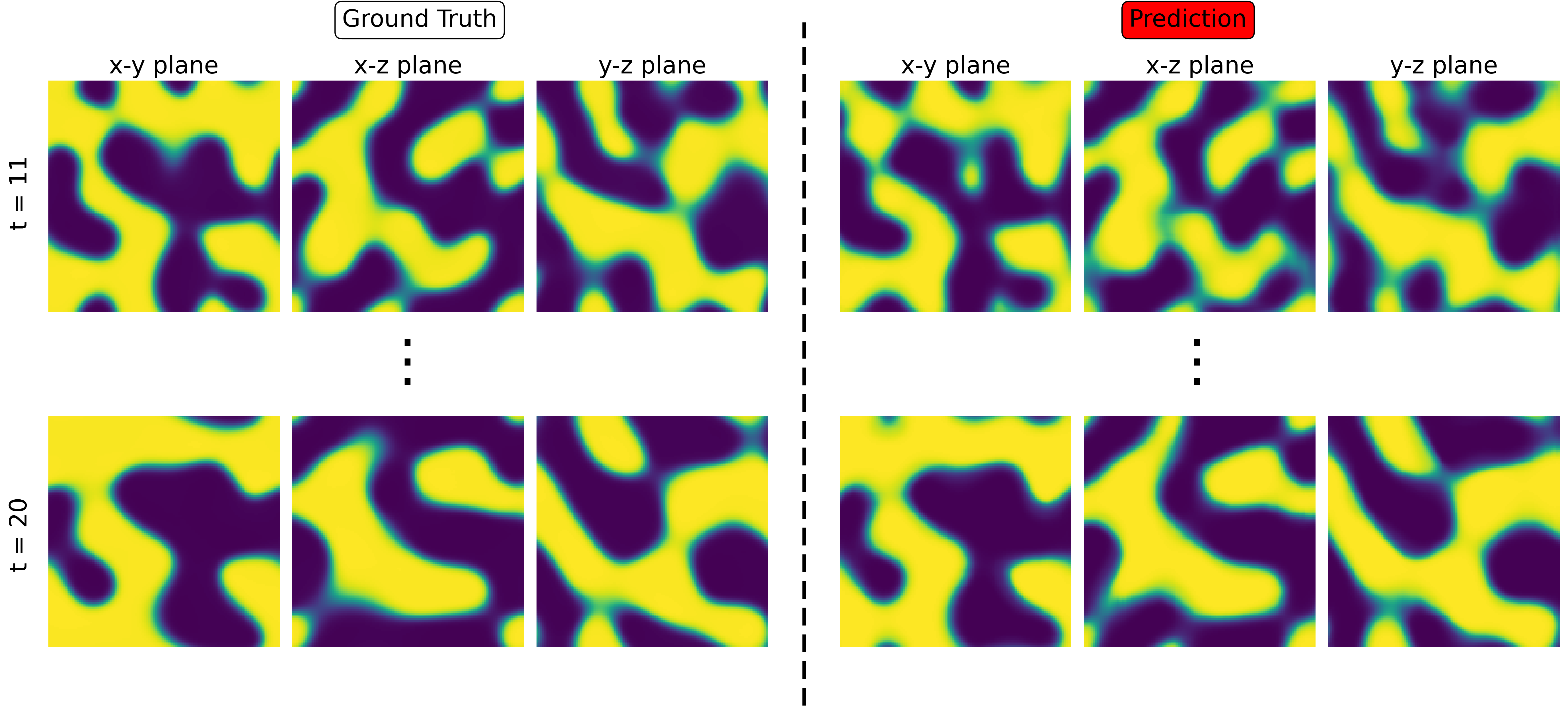}
\caption{\textbf{Spinodal decomposition prediction 2D slice visualization (10 input frames, 10 output frames):} Predictions from the proposed model are displayed alongside the ground-truth at timesteps $t = 11, 20$. For each timestep, central slices of the volume are shown in the axial, coronal, and sagittal planes.}
\label{fig6}
\end{figure}

For the $10\rightarrow10$ task, 370 test samples are evaluated. Training loss for this model is presented in Fig.~\ref{epochLoss}. A representative example is shown in Fig.~\ref{fig5}, which displays predicted frames at timestamps $t = 11, 14, 17,$ and $20$ for the proposed model. The corresponding input sequence is provided in Fig.~\ref{fig4} at timestamps $t = 1$ and $9$. In Fig.~\ref{fig5}, the column to the right of the prediction sequences reports the RMSE, 3D SSIM, and PSNR values for the proposed model, computed at each timestep. The interface curvature is shown in the bottom row at timestamps $t = 14$ and $20$, where the predicted values are directly compared with the ground-truth. To quantitatively assess the interface curvature distribution (ICD) plots beyond visual inspection, the total variation distance is computed between the ground-truth and predicted sequences; lower values indicate closer agreement with the ground-truth. 2D slices of both the ground-truth and predicted volumes are shown in Fig.~\ref{fig6}, taken from the center of the volume along the axial, coronal, and sagittal planes.

The accumulation of prediction error stabilizes as training progresses, indicating that the optimization process converges toward a stable solution that captures the dominant patterns of the training dataset. The predictive results of the proposed model closely match the ground-truth. Quantitatively, the model achieves an RMSE of approximately 0.007, a 3D SSIM exceeding 0.98, and a PSNR of 43 by timestep 20. These metrics indicate a high degree of visual similarity between the predicted and ground-truth sequences, which is further corroborated by the 2D slice visualizations shown in Fig.~\ref{fig6}. The interface curvature distribution of the predictions exhibits a shape comparable to that of the ground-truth, although noticeable discrepancies remain. When full input context is available, these results suggest that the proposed model is capable of accurately predicting three-dimensional microstructure evolution without explicit physics-based guidance. However, alternative prediction scenarios must be evaluated to more comprehensively assess the robustness and generalization capability of the model.

\begin{figure}[h!]
\centering
\includegraphics[width=0.9\linewidth]{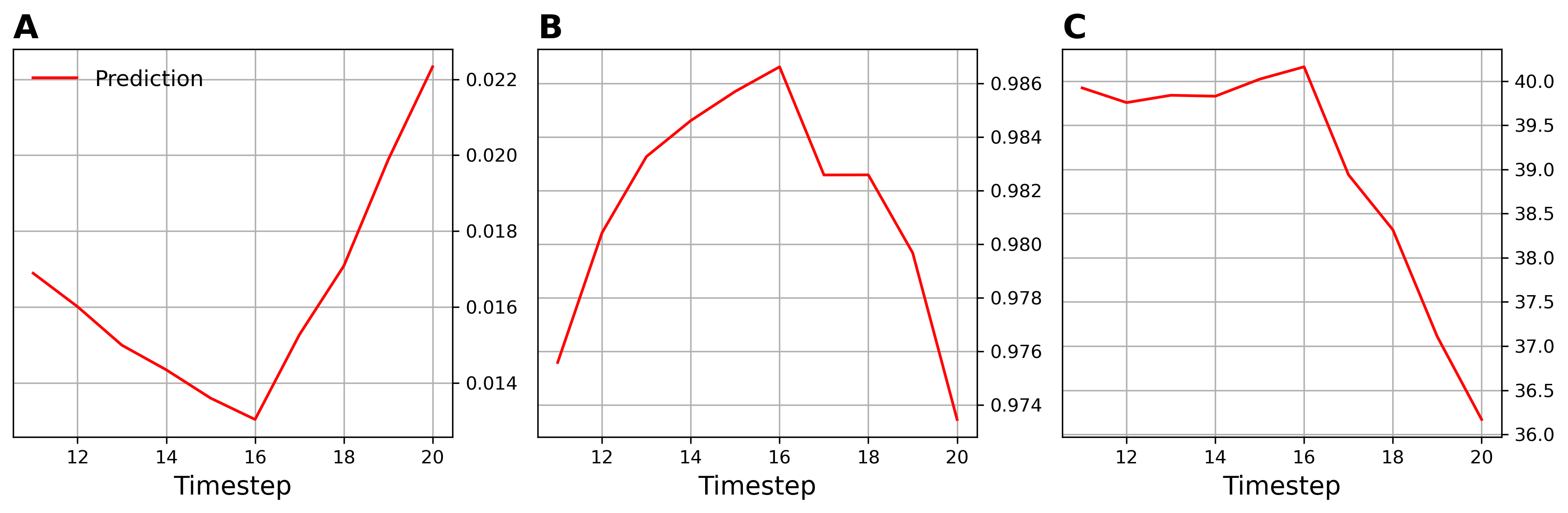}
\caption{\textbf{Spinodal decomposition average visual metrics (10 input frames, 10 output frames):} Dataset-averaged RMSE (\textbf{A}), 3D SSIM (\textbf{B}), and PSNR (\textbf{C}), evaluated at each timestep over the prediction horizon across all 370 samples in the experiment.}
\label{fig7}
\end{figure}

The average visual metric results are presented in Fig.~\ref{fig7}, summarizing the model’s performance across all 370 test samples. The average RMSE (Fig.~\ref{fig7}A) increases from approximately 0.017 at timestep 11 to just above 0.022 at timestep 20, reflecting the expected accumulation of error in high-resolution image prediction tasks. The average 3D SSIM (Fig.~\ref{fig7}B) remains consistently above 0.97 across all timesteps, indicating that the proposed model preserves strong structural similarity to the ground-truth sequences. The PSNR (Fig.~\ref{fig7}C) decreases from approximately 40 to around 36 over the prediction horizon, suggesting a gradual reduction in reconstruction fidelity while maintaining overall image quality. Notably, all three metrics exhibit a similar trend, with performance beginning to degrade more noticeably after approximately timestep 16. In the following experiments, we further investigate predictive fidelity by incorporating physics-guided training into the model.

\subsection{Predictions of 40 future frames from 10 input frames (10$\rightarrow$40)}
\label{subsec:10-40}

\begin{figure}[h!]
\centering
\includegraphics[width=0.45\linewidth]{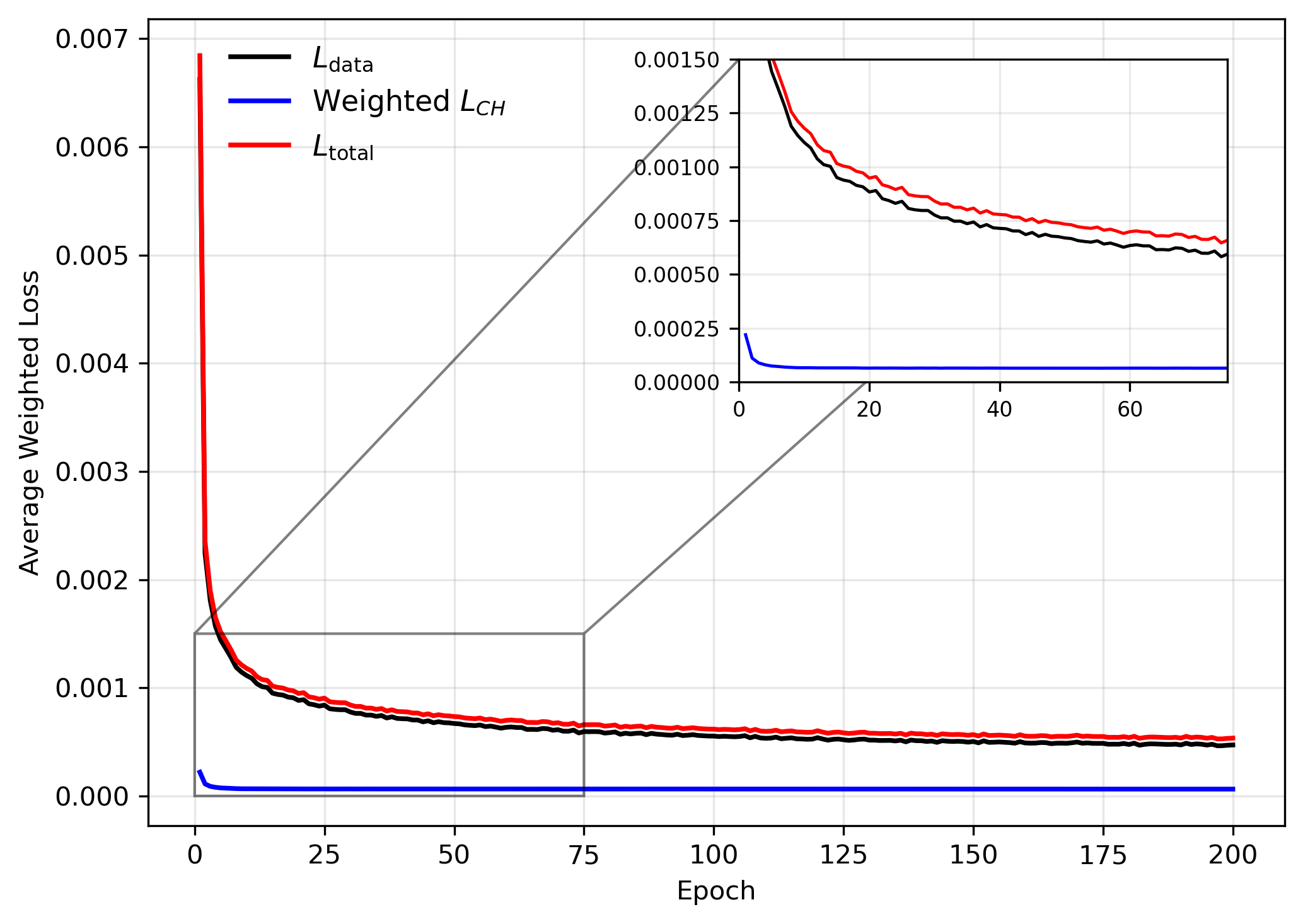}
\caption{\textbf{Physics-guided model total training losses:} Loss contribution to model training is reported at every epoch, illustrating the contribution of data loss and physics-guided loss. Physics guidance hs a regularizer value of \(\lambda_{phy} = 10^{-8}\).}
\label{epochLossPhy}
\end{figure}

\begin{figure}[h!]
\centering
\includegraphics[width=0.45\linewidth]{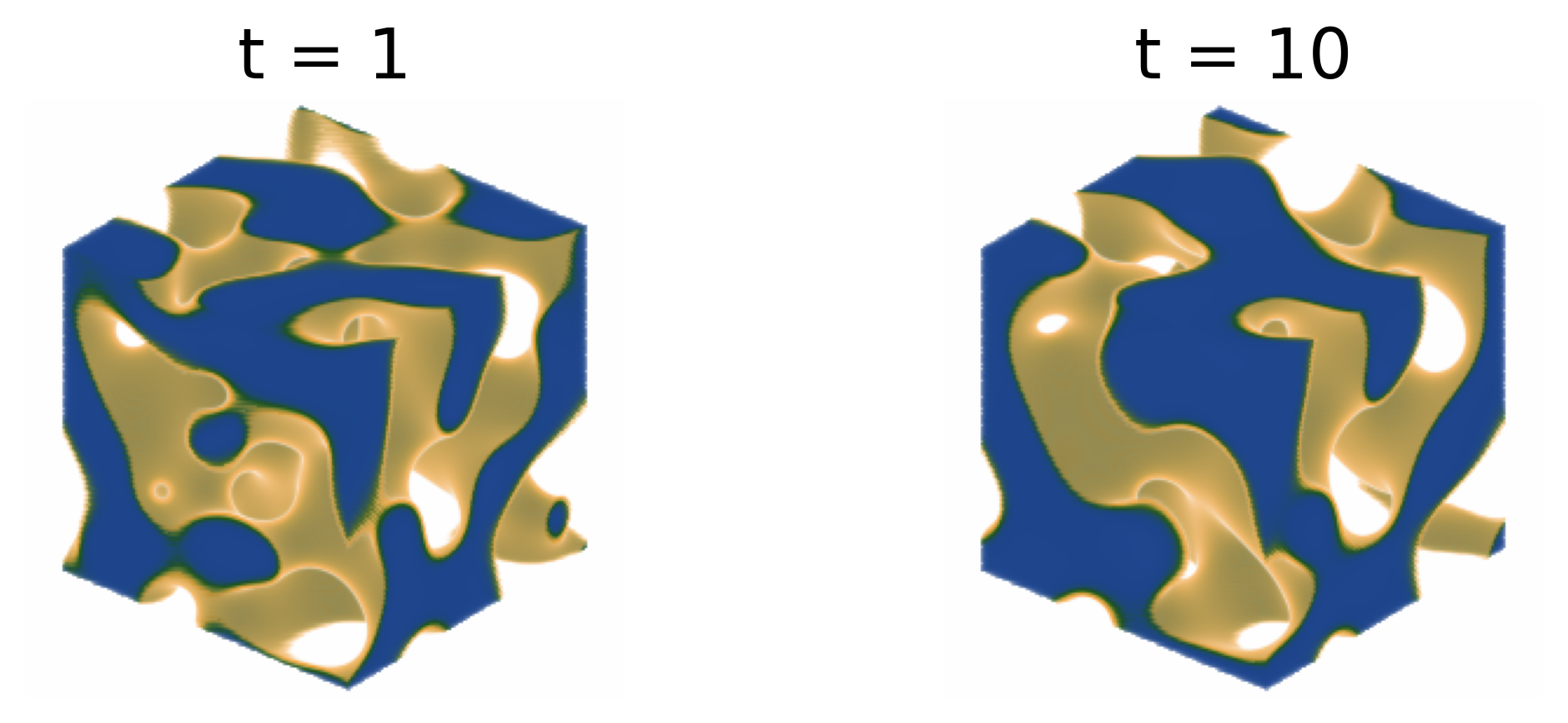}
\caption{\textbf{Spinodal decomposition input (10 input frames, 40 output frames):} Input sequence corresponding to the results shown in Fig.~\ref{fig13}, displayed at timesteps $t = 1, 10$.}
\label{fig12}
\end{figure}

\begin{figure}[h!]
\centering
\includegraphics[width=\linewidth]{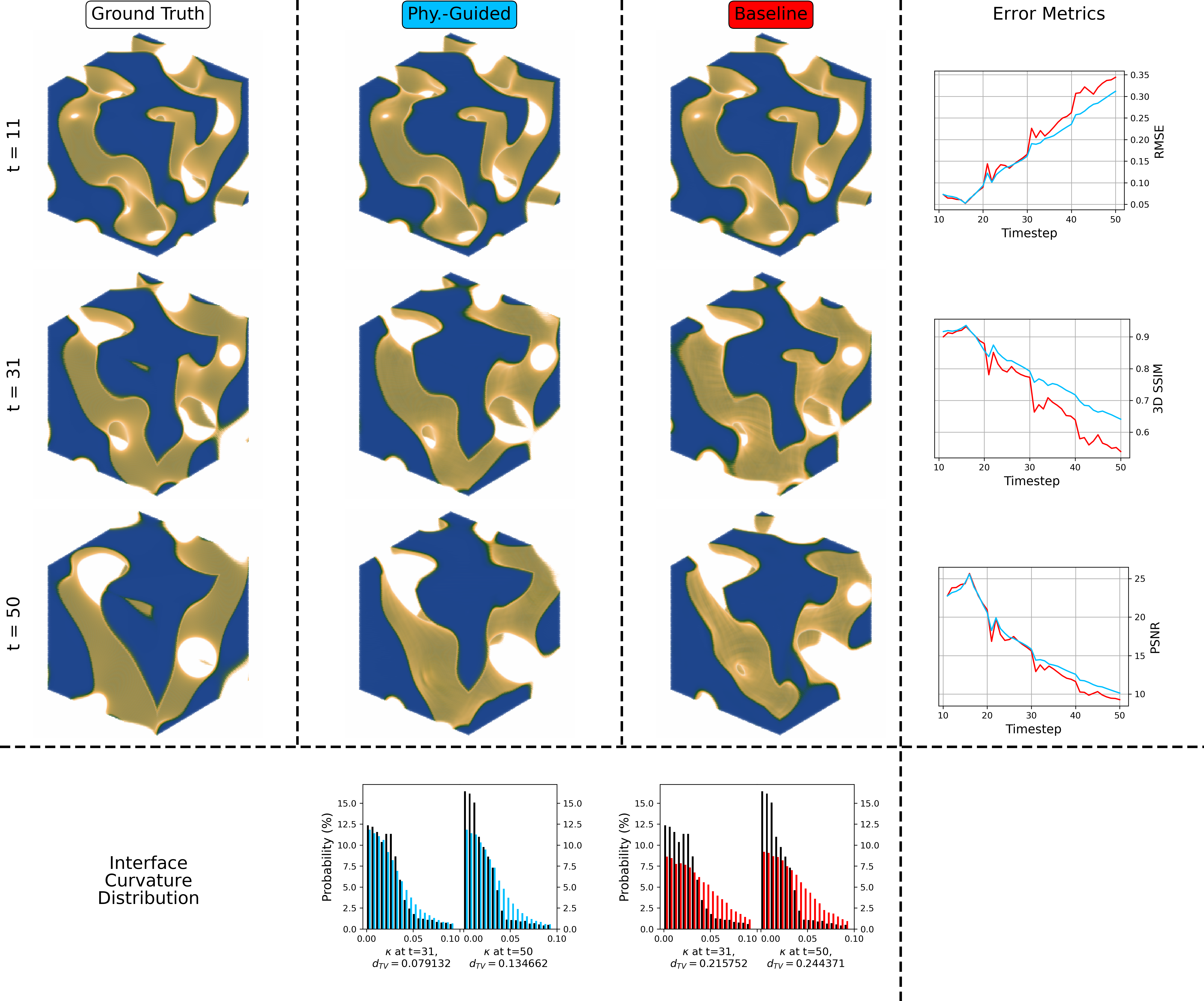}
\caption{\textbf{Spinodal decomposition prediction (10 input frames, 40 output frames) comparison:} Predictions from the physics-guided and baseline models are shown alongside the ground-truth at timesteps $t = 11, 31, 50$. The right column reports the RMSE, 3D SSIM, and PSNR computed between the predicted and ground-truth sequences for both models at each timestep. The bottom row displays the interface curvature distributions for each model compared with the ground-truth at $t = 31, 50$, with the corresponding total variation distance provided beneath each distribution.}
\label{fig13}
\end{figure}

\begin{figure}[h!]
\centering
\includegraphics[width=\linewidth]{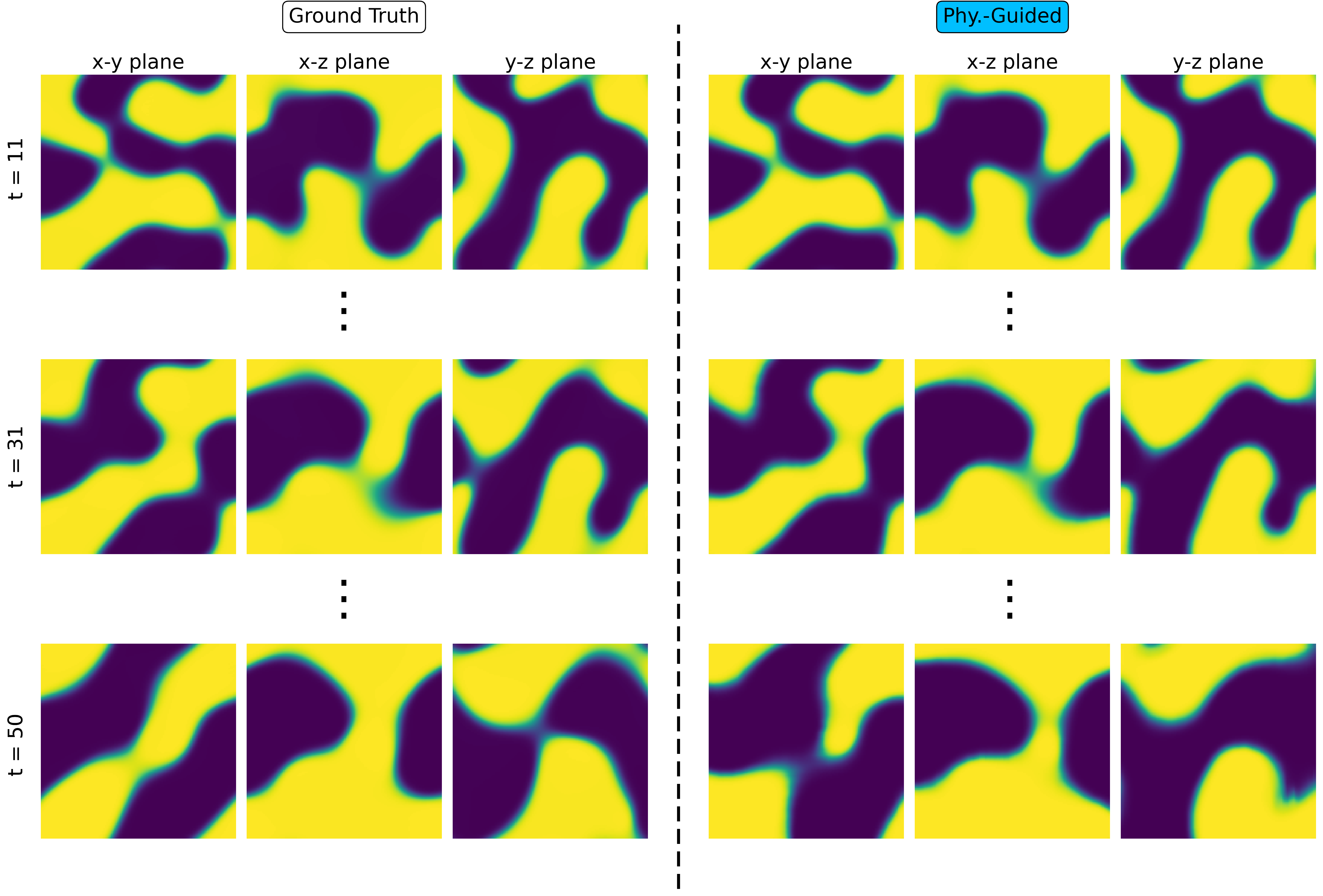}
\caption{\textbf{Spinodal decomposition prediction 2D slice visualization (10 input frames, 40 output frames):} Predictions from the physics-guided model are displayed alongside the ground-truth at timesteps $t = 11, 31, 50$. For each timestep, central slices of the volume are shown in the axial, coronal, and sagittal planes.}
\label{fig14}
\end{figure}

For the $10 \rightarrow 40$ task, 270 test samples are evaluated. Training loss for this model is presented in Fig.~\ref{epochLossPhy}. One representative example is shown in Fig.~\ref{fig13}, which displays predicted frames at timestamps $t = 11, 31, 50$ for both the baseline and physics-guided models, and the input for the experiment is provided in Fig.~\ref{fig12} at timestamps $t = 1, 10$. The column to the right of the prediction sequences reports the RMSE, 3D SSIM, and PSNR values for both models, computed at each frame. Below each prediction sequence, the interface curvature is shown at timestamps $t = 31, 50$, where the predicted values are directly compared with the ground-truth. To quantitatively assess the ICD plots beyond visual inspection, the total variation distance is computed between the ground-truth and predicted sequences for each model; lower values indicate closer agreement with the ground-truth. 2D slices of both the ground-truth and predicted volumes are shown in Fig.~\ref{fig14}, taken from the center of the volume along the axial, coronal, and sagittal planes.

The error contributions shown in Fig.~\ref{epochLossPhy} demonstrate that the physics guidance contributes substantially less to the total loss than the data loss. Both the data and physics losses stabilize as training progresses, indicating that the physics-guided objective converges without introducing persistent instability during optimization. This behavior is particularly important for the incorporation of the physics guidance, as the relatively small \(\lambda_{\mathrm{phy}}\) is intentionally selected to prevent the physics-based regularization from over-constraining the data-driven objective. Consequently, the physics guidance serves as an auxiliary constraint that promotes physically consistent predictions while preserving the primary influence of the data loss.

In this extrapolative prediction task, both models generate sequences that resemble the ground truth; however, the baseline model exhibits more pronounced degradation over time. For the representative example shown in Fig.~\ref{fig13}, the physics-guided model yields favorable reconstruction metrics at several evaluated times, indicating improved voxel-wise agreement and structural similarity with the ground truth. The physics guidance improves voxel-level reconstruction in extrapolative prediction tasks and supports the preservation of key microstructural features. This distinction is evident in the 2D slice visualizations in Fig.~\ref{fig14}, where the physics-guided predictions remain structurally consistent with the ground truth despite visible discrepancies. Beyond image-based metrics, the physics-guided model produces an interface curvature distribution (ICD) that more closely aligns with the ground truth, further supported by a lower total variation distance at timestep 50. Overall, these results indicate that the physics-guided model better sustains accurate long-horizon forecasts by stabilizing predictions and preserving key microstructural characteristics.

\begin{figure}[h!]
\centering
\includegraphics[width=0.9\linewidth]{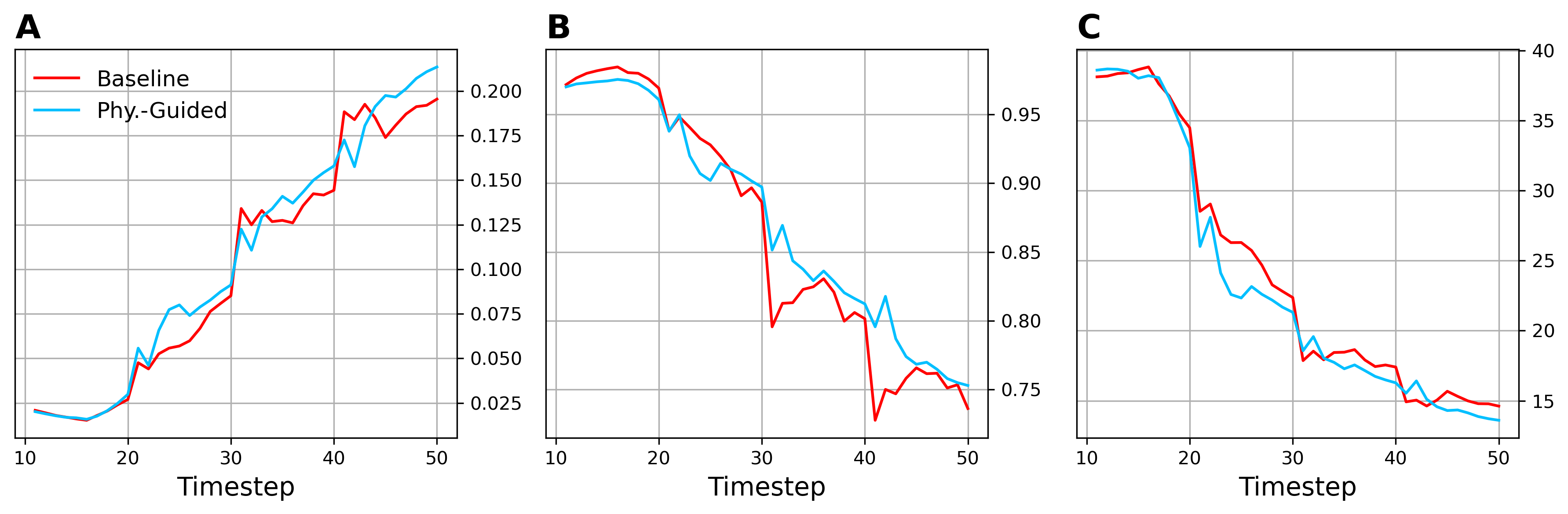}
\caption{\textbf{Spinodal decomposition average visual metrics (10 input frames, 40 output frames):} Dataset-averaged RMSE (\textbf{A}), 3D SSIM (\textbf{B}), and PSNR (\textbf{C}), evaluated at each timestep over the prediction horizon across all 270 samples in the experiment.}
\label{fig15}
\end{figure}

Figure~\ref{fig15} presents the average metric scores for both models across all 270 test samples. The results are similar in magnitude, particularly prior to timestep 20, and both models exhibit comparable error accumulation trends with only minor discrepancies. At timestep 50, the average RMSE (Fig.~\ref{fig15}A) is approximately 0.21 for the physics-guided model and 0.19 for the baseline model. At the same timestep, the average 3D SSIM (Fig.~\ref{fig15}B) and PSNR (Fig.~\ref{fig15}C) are approximately 0.75 and 14, respectively. Although the physics-guided model does not yield substantial improvements in standard visual metrics, it exhibits reduced degradation relative to the baseline model. More importantly, its predictions better preserve the underlying physical structure of the microstructure, as evidenced by the interface curvature distribution (ICD) in Fig.~\ref{fig13}. In the context of extrapolative forecasting, these results indicate that the physics-guided model improves physical fidelity by stabilizing microstructural evolution, rather than strictly enhancing voxel-wise accuracy.

\subsection{Predictions of 15 future frames from 5 input frames (5$\rightarrow$15)}

\begin{figure}[h!]
\centering
\includegraphics[width=0.45\linewidth]{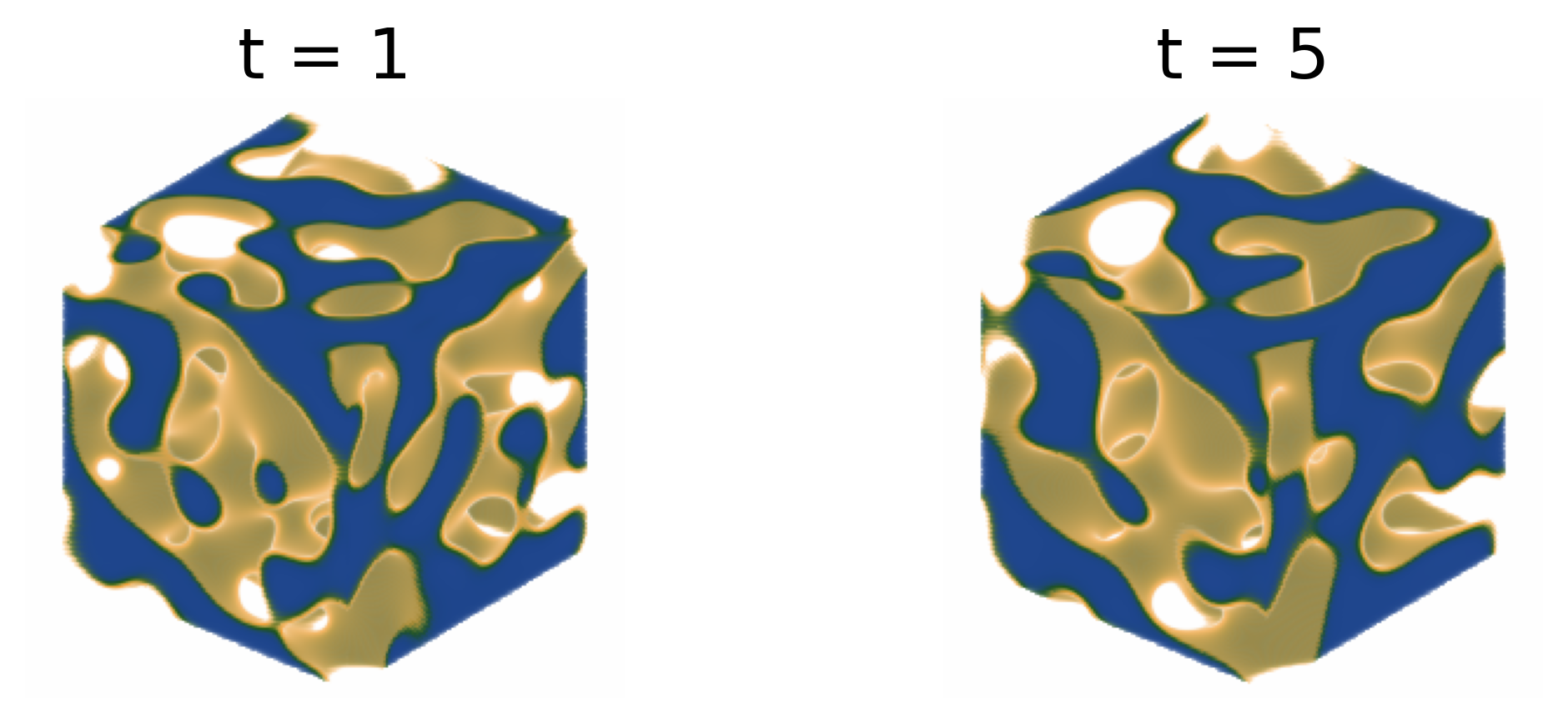}
\caption{\textbf{Spinodal decomposition input (5 input frames):} Input sequence corresponding to the results shown in Fig.~\ref{fig9}, displayed at timesteps $t = 1, 5$.}
\label{fig8}
\end{figure}

\begin{figure}[h!]
\centering
\includegraphics[width=\linewidth]{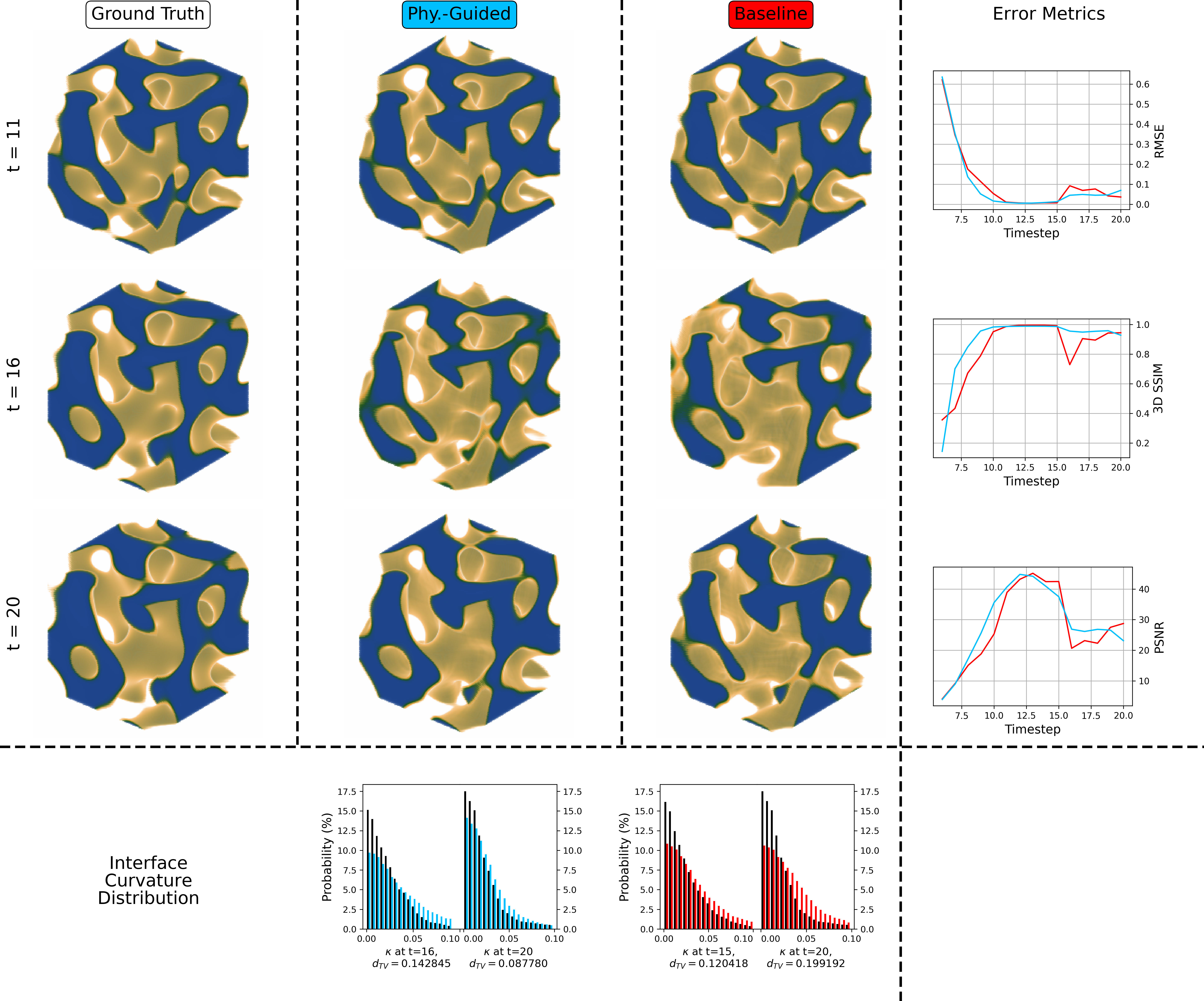}
\caption{\textbf{Spinodal decomposition prediction (5 input frames, 15 output frames) comparison:} Predictions from the physics-guided and baseline models are shown alongside the ground-truth at timesteps $t = 11, 16, 20$. The right column reports the RMSE, 3D SSIM, and PSNR computed between the predicted and ground-truth sequences for both models at each timestep. The bottom row displays the interface curvature distributions for each model compared with the ground-truth at $t = 16, 20$, with the corresponding total variation distance provided beneath each distribution.}
\label{fig9}
\end{figure}

\begin{figure}[h!]
\centering
\includegraphics[width=\linewidth]{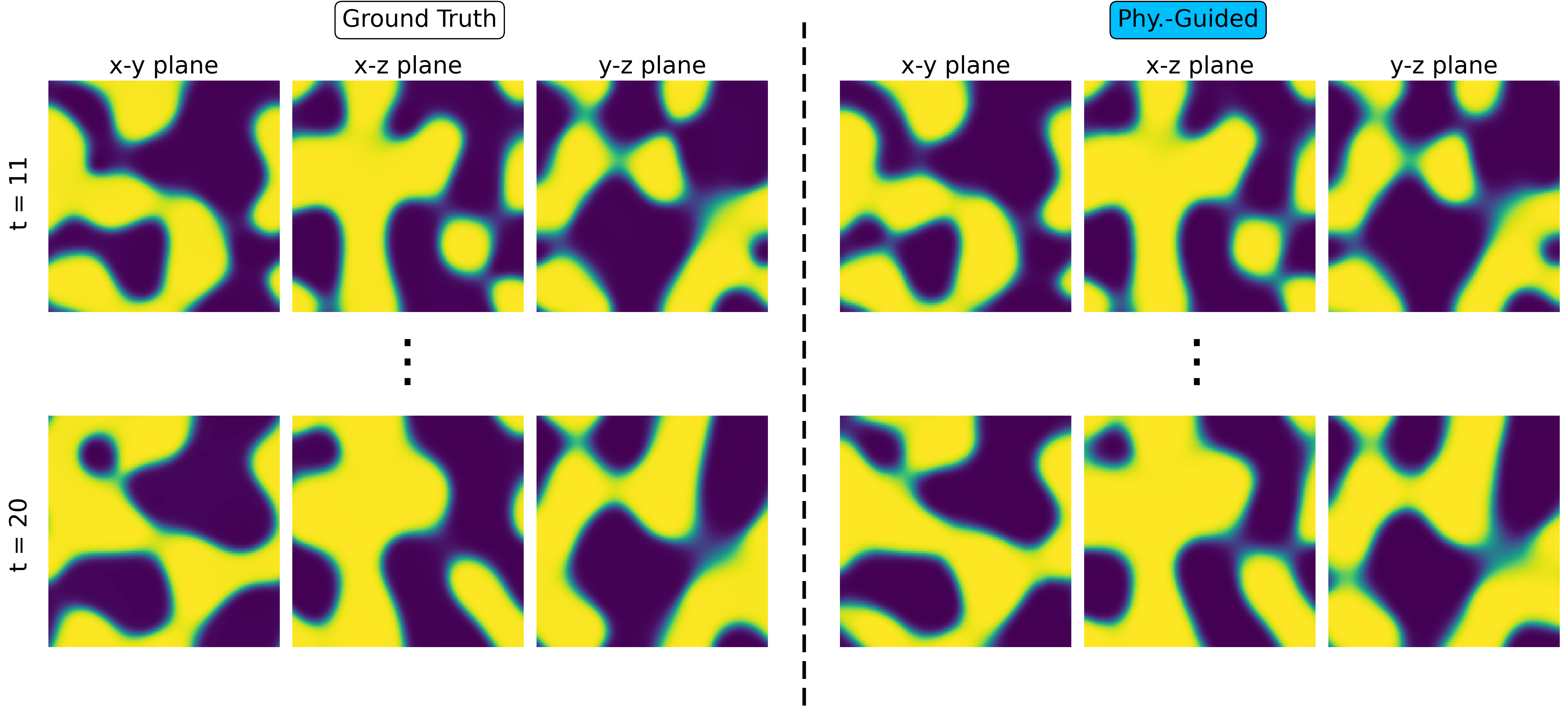}
\caption{\textbf{Spinodal decomposition prediction 2D slice visualization (5 input frames, 15 output frames):} Predictions from the physics-guided model are displayed alongside the ground-truth at timesteps $t = 11, 20$. For each timestep, central slices of the volume are shown in the axial, coronal, and sagittal planes.}
\label{fig10}
\end{figure}

For the $5 \rightarrow 15$ task, 370 test samples are evaluated. One representative example is shown in Fig.~\ref{fig9}, which displays predicted frames at timestamps $t = 11, 16, 20$ for both the baseline and physics-guided models, and the input for the experiment is provided in Fig.~\ref{fig8} at timestamps $t = 1, 5$. The column to the right of the prediction sequences reports the RMSE, 3D SSIM, and PSNR values for both models, computed at each frame. Below each prediction sequence, the interface curvature is shown at timestamps $t = 16, 20$, where the predicted values are directly compared with the ground-truth. To quantitatively assess the ICD plots beyond visual inspection, the total variation distance is computed between the ground-truth and predicted sequences for each model; lower values indicate closer agreement with the ground-truth. 2D slices of both the ground-truth and predicted volumes are shown in Fig.~\ref{fig10}, taken from the center of the volume along the axial, coronal, and sagittal planes.

Both models closely reproduce the bicontinuous morphology of the ground-truth. In this prediction scenario, however, the physics-guided model exhibits greater structural robustness and morphology fidelity relative to the baseline model. The physics-guided model demonstrates more consistent performance across timesteps and mitigates the sharp error accumulation observed in the baseline model. The 2D slice visualizations shown in Fig.~\ref{fig10} further demonstrate the predictive fidelity of the physics-guided model. At timestep 20, the predicted sequence closely matches the ground-truth across all three slices, indicating strong internal structural agreement beyond 3D morphology. In addition to these improvements, the interface curvature distribution of the physics-guided predictions is significantly closer to that of the ground-truth than the baseline model. This is particularly evident when comparing the total variation distances between the predicted and ground-truth distributions for each model. At timestep 20, the total variation distance of the physics-guided model is less than half that of the baseline model. Under reduced input conditions such as this scenario, these results indicate that physics guidance enhances predictive robustness. This is particularly important in scenarios where a full microstructure sequence may not be readily available for the nominal input of 10 frames.

\begin{figure}[h!]
\centering
\includegraphics[width=0.9\linewidth]{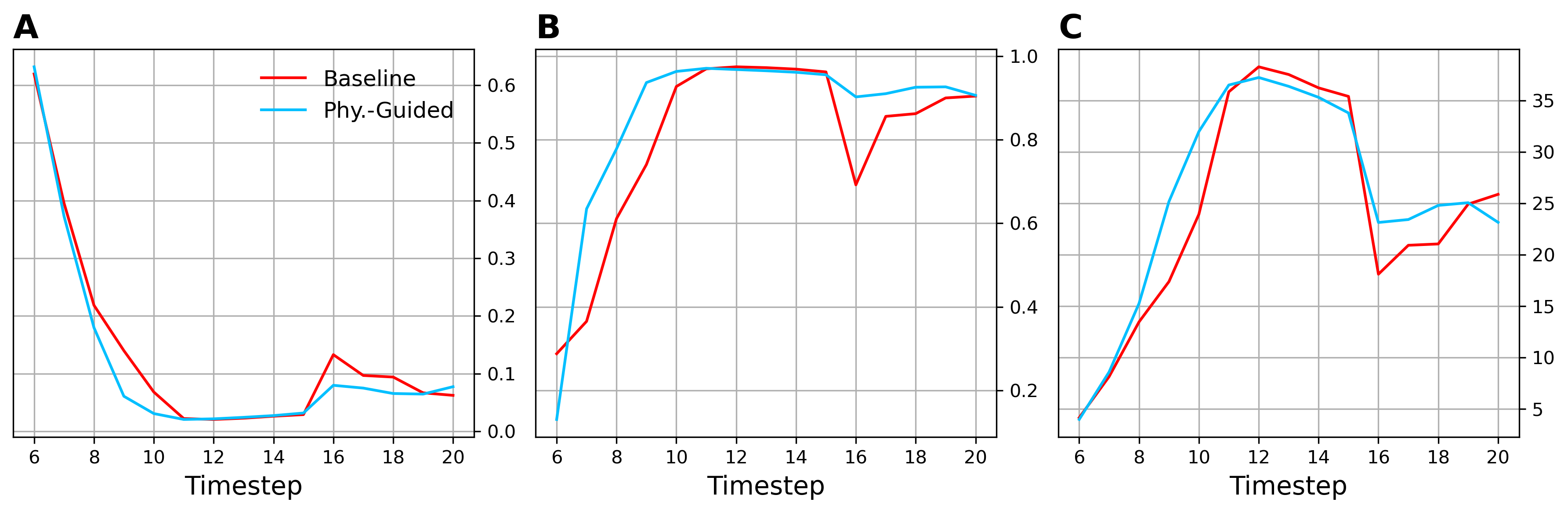}
\caption{\textbf{Spinodal decomposition average visual metrics (5 input frames, 15 output frames):} Dataset-averaged RMSE (\textbf{A}), 3D SSIM (\textbf{B}), and PSNR (\textbf{C}), evaluated at each timestep over the prediction horizon across all 370 samples in the experiment.}
\label{fig11}
\end{figure}

The average visual metric results are presented in Fig.~\ref{fig11}, summarizing both models’ performance across all 370 test samples. Both models exhibit similar trends across the visual metrics, with only minor differences at specific timesteps. At timestep 20, the average RMSE (Fig.~\ref{fig11}A) is approximately 0.065, the average 3D SSIM (Fig.~\ref{fig11}B) is around 0.9, and the average PSNR (Fig.~\ref{fig11}C) is approximately 26 for the baseline model and 24 for the physics-guided model. All three metrics show elevated error prior to frame 10, which is attributed to the use of zero-padding in the input sequence. However, both models recover from these short-term artifacts and produce higher-quality predictions at later timesteps. Consistent with the trends observed in Fig.~\ref{fig9}, the physics-guided model does not exhibit the same sharp error increase at timestep 16 as the baseline model. Additionally, it recovers more rapidly during the early prediction frames. These results suggest that the physics-guided model maintains higher predictive fidelity under reduced-input conditions compared to the baseline model.

\subsection{Predictions of 19 future frames from 1 input frame (1$\rightarrow$19)}

\begin{figure}[h!]
\centering
\includegraphics[width=0.25\linewidth]{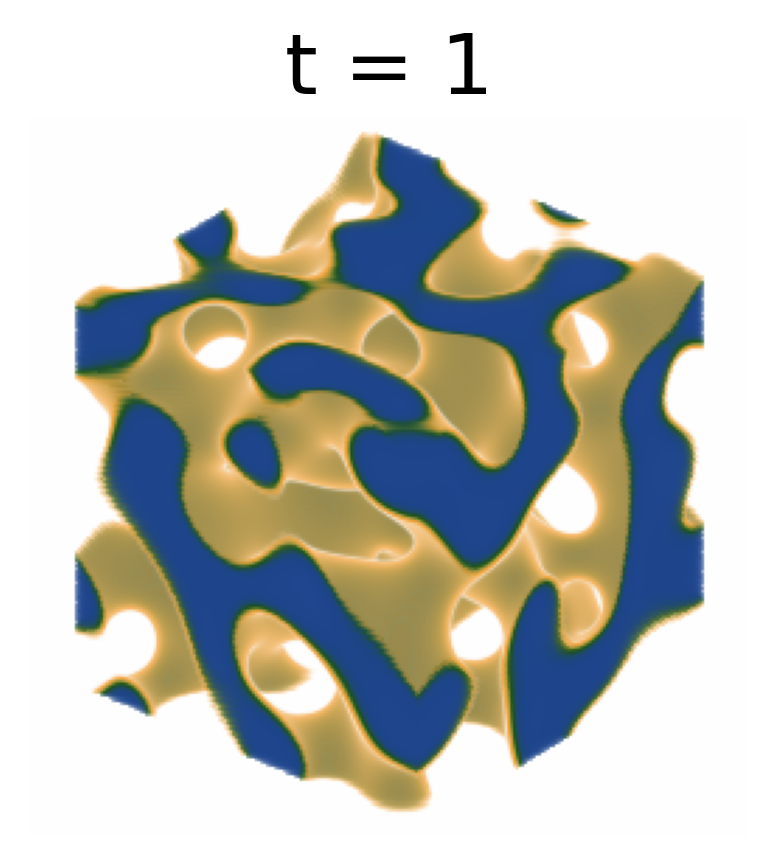}
\caption{\textbf{Spinodal decomposition input (1 input frame):} Input frame corresponding to the results shown in Fig.~\ref{fig17}, displayed at timestep $t = 1$.}
\label{fig16}
\end{figure}

\begin{figure}[h!]
\centering
\includegraphics[width=\linewidth]{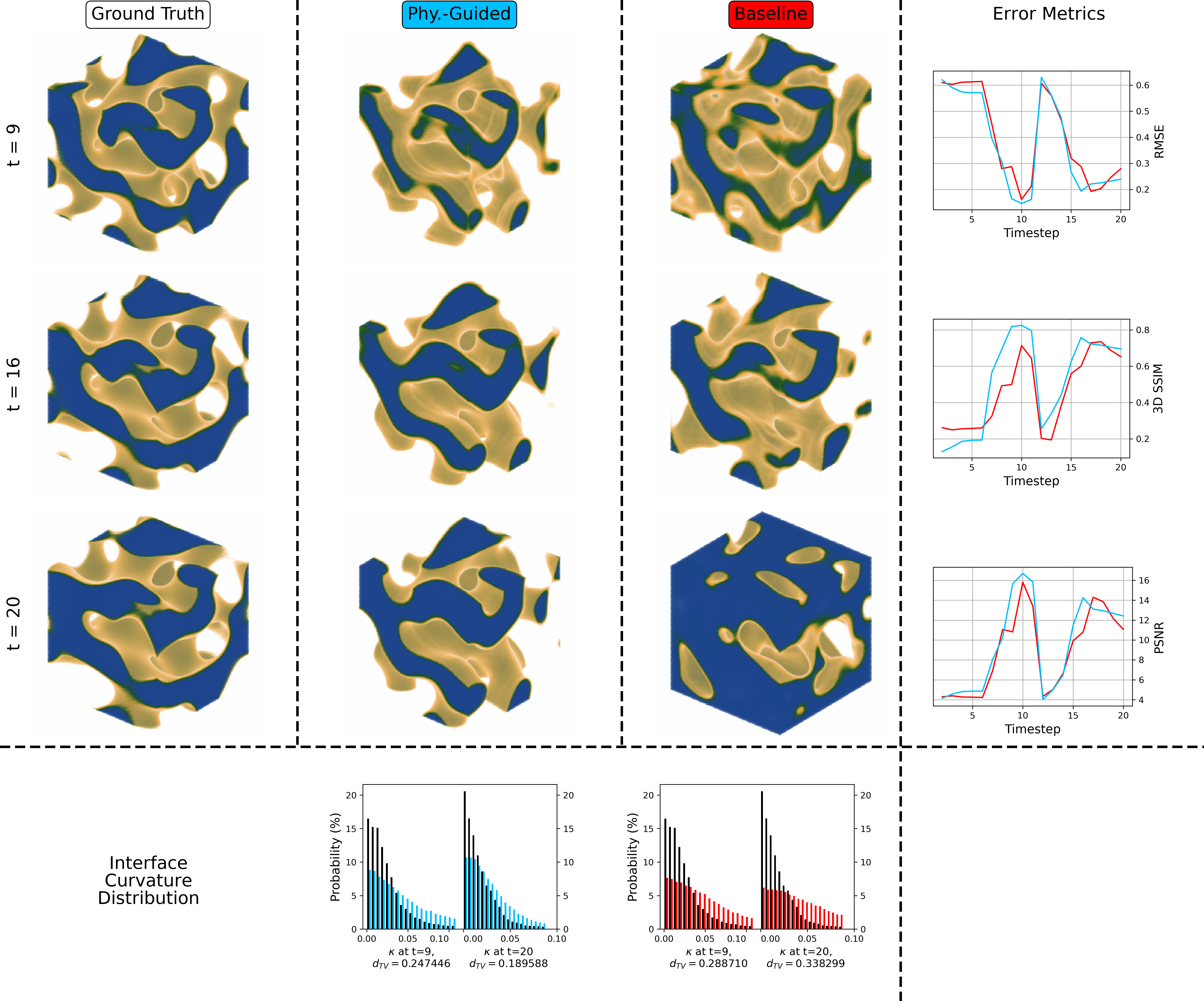}
\caption{\textbf{Spinodal decomposition prediction (19 output frames) comparison:} Predictions from the physics-guided and baseline models are shown alongside the ground-truth at timesteps $t = 9, 16,$ and $20$. The right column reports the RMSE, 3D SSIM, and PSNR computed between the predicted and ground-truth sequences for both models at each timestep. The bottom row displays the interface curvature distributions for each model compared with the ground-truth at $t = 9, 20$, with the corresponding total variation distance provided beneath each distribution.}
\label{fig17}
\end{figure}

\begin{figure}[h!]
\centering
\includegraphics[width=\linewidth]{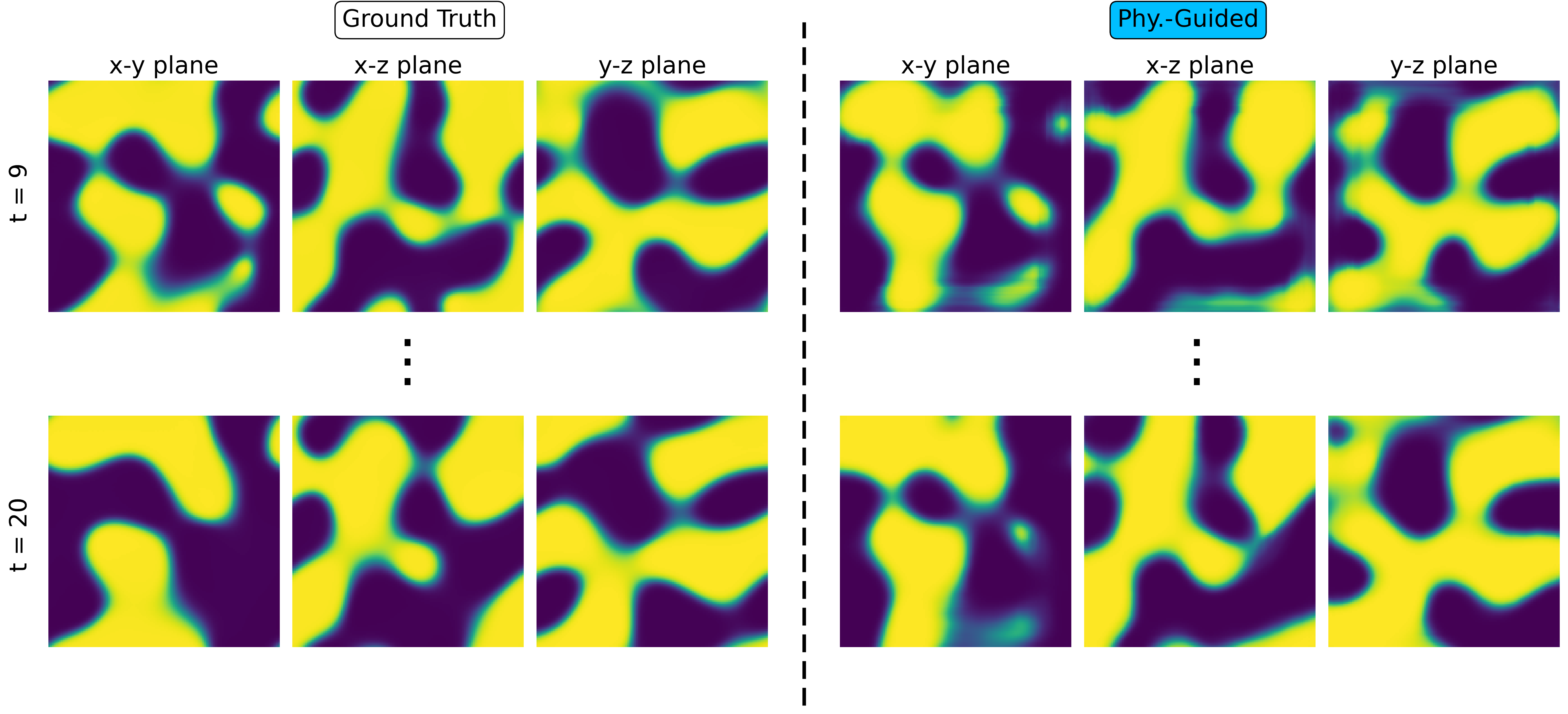}
\caption{\textbf{Spinodal decomposition prediction 2D slice visualization (1 input frame, 19 output frames):} Predictions from the physics-guided model are displayed alongside the ground-truth at timesteps $t = 9, 20$. For each timestep, central slices of the volume are shown in the axial, coronal, and sagittal planes.}
\label{fig18}
\end{figure}

For the $1 \rightarrow 19$ task, 370 test samples are evaluated. One representative example is shown in Fig.~\ref{fig17}, which displays predicted frames at timestamps $t = 9, 16, 20$ for both the baseline and physics-guided models, and the input for the experiment is provided in Fig.~\ref{fig16} at timestamp $t = 1$. The column to the right of the prediction sequences reports the RMSE, 3D SSIM, and PSNR values for both models, computed at each frame. Below each prediction sequence, the interface curvature is shown at timestamps $t = 9, 20$, where the predicted values are directly compared with the ground-truth. To quantitatively assess the ICD plots beyond visual inspection, the total variation distance is computed between the ground-truth and predicted sequences for each model; lower values indicate closer agreement with the ground-truth. Two-dimensional slices of both the ground-truth and predicted volumes are shown in Fig.~\ref{fig18}, taken from the center of the volume along the axial, coronal, and sagittal planes.

Unlike the previous experiments, the temporal context in the input is limited to a single frame, requiring both models to extrapolate future states. Under this constraint, the physics-guided model produces more informed predictions than the baseline model. Although both models yield similar visual metric scores, the predicted sequences indicate that the physics-guided model achieves substantially higher fidelity relative to the baseline. Fig.~\ref{fig18} presents 2D slice visualizations of the physics-guided predictions alongside the ground-truth. While noticeable discrepancies remain at the evaluated timesteps, the predicted microstructures retain strong qualitative similarity. In addition to visual agreement, the physics-guided model provides a more accurate representation of the ICD. This is evident not only in the reduced total variation distance, but also in the closer alignment between the predicted and ground-truth distributions. Overall, these results highlight the benefit of incorporating physics guidance in spatiotemporal forecasting, particularly in low-context regimes where purely data-driven models struggle to maintain physically consistent predictions.

\begin{figure}[h!]
\centering
\includegraphics[width=0.9\linewidth]{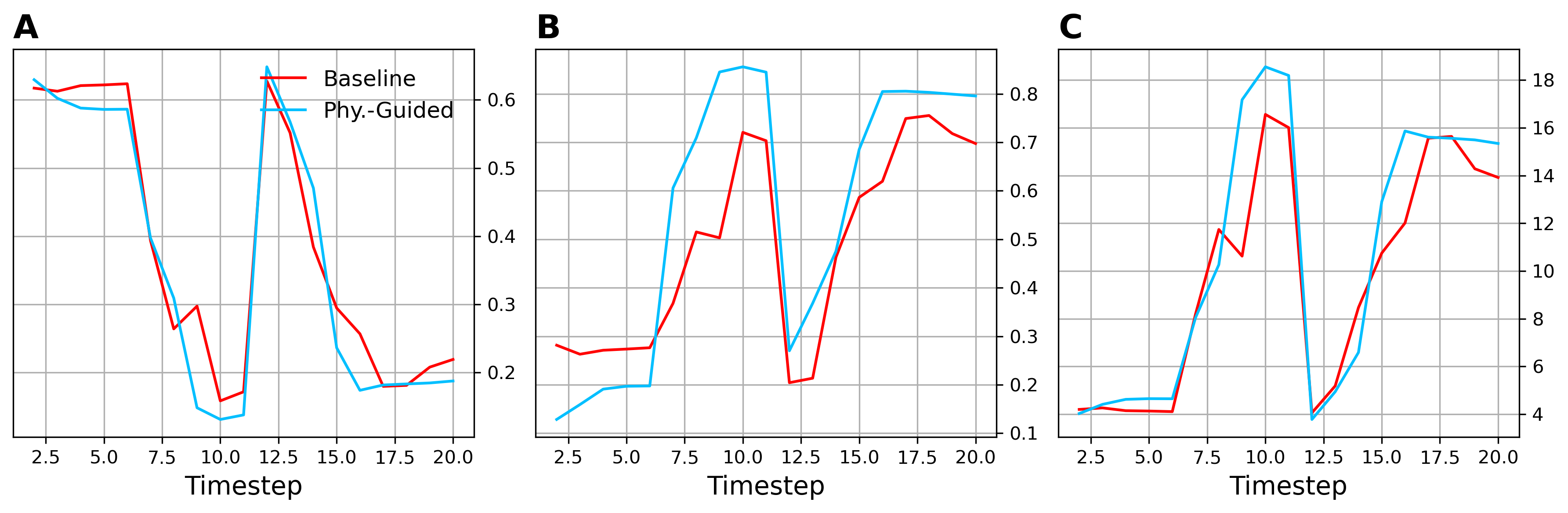}
\caption{\textbf{Spinodal decomposition average visual metrics (1 input frame, 19 output frames):} Dataset-averaged RMSE (\textbf{A}), 3D SSIM (\textbf{B}), and PSNR (\textbf{C}), evaluated at each timestep over the prediction horizon across all 370 samples in the experiment.}
\label{fig19}
\end{figure}

Figure~\ref{fig19} presents the average metric scores for both models across all 370 test samples. All metrics exhibit noticeable fluctuations over time for both models. Despite these variations, each model produces coherent microstructure evolution without being dominated by noise. The average RMSE (Fig.~\ref{fig19}A) for both models ranges from approximately 0.10 to 0.65. The average 3D SSIM (Fig.~\ref{fig19}B) is consistently higher for the physics-guided model across most timesteps, with a notable difference at timestep 20 (0.8 for the physics-guided model versus 0.7 for the baseline). A similar trend is observed in the average PSNR (Fig.~\ref{fig19}C), where the physics-guided model achieves a value of approximately 15.5 at timestep 20, compared to about 14 for the baseline model. Taken together with the preceding results in this experiment, these findings indicate that the physics-guided model is more robust under limited-input conditions for microstructure forecasting.

\subsection{Computational Efficiency and Inference Speedup}
The proposed model and physics-guided model exhibit a difference in training time of approximately 2 hours. However, their inference times are equivalent when generating 10-frame prediction sequences. Because the physics-guided loss is applied only during training and does not contribute to the inference forward-pass, physics guidance introduces no additional inference-time cost. Compared with \texttt{SpectralETD}, both deep learning models use approximately \(32.3\times\) less computational time to generate corresponding microstructure evolution at a spatial resolution \(128^3\). This significant computational advantage is beneficial for high-throughput applications, where the initial training cost can be amortized across repeated predictions. All simulation and inferences are run on the same hardware as reported in Sec.~\ref{netArch}.

\begin{table}[h!]
\centering
\begin{tabular}{|c|c|c|c|c|}
\hline
\textbf{\makecell{Models \\ (200 Epochs, \\ batch size 1)}} & \textbf{\makecell{Training Time \\ (hours)}} & \textbf{\makecell{Average \\ Inference \\ Time \\ (40 frames \(64^3\))}} &
\textbf{\makecell{Average \\ Inference \\ Time \\ (40 frames \(96^3\))}} &
\textbf{\makecell{Average \\ Inference \\ Time \\ (40 frames \(128^3\))}} \\
\hline
The Proposed Model & 46.1561 & 0.0582 & 0.1823 & 0.3682 \\
Physics-Guided Model & 48.4778 & 0.0582 & 0.1823 & 0.3682 \\
SpectralETD & X & 9.0483 & 9.129 & 11.896 \\
\hline
\end{tabular}
\caption{\textbf{Inference efficiency comparison across deep spatiotemporal models and computational models on spinodal decomposition prediction}. Training time and inference time for 3D spinodal decomposition prediction.}
\label{tab:runtime}
\end{table}

\section{Discussion}
\label{sec:disc}
The present results demonstrate that fully convolutional
spatiotemporal learning can be extended to dense 3D microstructure
evolution while retaining both high predictive fidelity and very low
inference cost. Building on the encoder--translator--decoder paradigm
of SimVP/SimVPv2, the proposed framework reformulates the spatial and
latent representations for volumetric phase-field data. In particular,
the factorized translator separates temporal evolution, local 3D
spatial interactions, and channel-wise feature coupling. This
formulation enables the network to learn a finite-horizon volumetric
evolution operator that maps an observed sequence directly to multiple
consecutive future 3D states within each prediction block. The model
therefore differs conceptually from surrogate formulations that advance
the system through a sequence of one-step state updates.

The nominal prediction results show that the proposed 3D formulation
can accurately learn Cahn--Hilliard microstructure evolution even
without explicit physics regularization. With sufficient temporal
context, the data-driven model reproduces the evolving bicontinuous
morphology with high structural fidelity over the complete prediction
block. This result is important because it indicates that the
factorized latent representation is capable of preserving both
volumetric morphology and temporal progression despite the high
dimensionality of the underlying field. The orthogonal 2D sections
provide additional evidence that the agreement is not limited to the
rendered surface morphology, but also extends to internal structures
within the predicted volumes.

A central finding of this study is that the value of physics guidance
depends strongly on the amount of temporal information available to
the predictor. When the full input sequence is provided, the observed
history already constrains the subsequent evolution strongly, and the
data-driven model achieves high accuracy. Under extended block-wise
forecasting, the benefit of the CH residual is reflected
more clearly in structural stability and interface-level morphology
than in uniformly improved voxel-wise reconstruction metrics. This
distinction is consistent with the long-horizon results, where the two
models exhibit comparable overall reconstruction trends while the
physics-guided predictions better preserve representative
interface-curvature distributions.

The importance of physics guidance becomes increasingly evident as the
available temporal context is reduced. With only five observed frames,
both models remain capable of reconstructing the dominant
microstructural morphology, but the physics-guided predictions exhibit
greater structural stability during portions of the forecast where the
data-driven model undergoes sharper degradation. With only a single
observed frame, the separation between the two formulations becomes
more pronounced: the physics-guided model maintains stronger
structural similarity over much of the prediction horizon and produces
interface-curvature statistics that more closely follow the reference
evolution. Taken together, these experiments suggest a clear
information-dependent role for the physical prior: as temporal
observations provide less direct information about the underlying
dynamics, equation-based regularization becomes increasingly important
for constraining the learned evolution.

This behavior also highlights an important distinction between
voxel-wise reconstruction accuracy and morphology-level fidelity.
Metrics such as RMSE, SSIM, and PSNR measure agreement with a specific
reference realization, whereas interface-curvature distributions
characterize geometric properties of the evolving phase boundaries.
Small displacements of an interface can produce noticeable voxel-wise
errors while leaving important aspects of the morphology and
coarsening behavior comparatively well preserved. Conversely, a model
with favorable image-level metrics may still exhibit undesirable
changes in interface geometry during extended forecasting. For this
reason, evaluating learned microstructure evolution using both
reconstruction metrics and morphology-sensitive quantities provides a
more informative assessment than relying on image similarity alone.

Computational efficiency is another central outcome of the proposed
framework. Because physics guidance is introduced only through the
training objective, the data-driven and physics-guided models share the
same inference pathway and therefore have identical deployment cost.
The resulting volumetric prediction is highly efficient across all
tested spatial resolutions. At the largest resolution considered,
the framework generates a complete sequence of 40 consecutive
\(128^3\) fields in 0.3682~s, corresponding to approximately
9.2~ms per volumetric state. Under the same reported computational
setting, the corresponding \texttt{SpectralETD} evolution requires 11.896~s,
yielding an approximately \(32.3\times\) wall-clock speedup. The
inference time also increases favorably as the spatial grid is enlarged
from \(64^3\) to \(128^3\). These results show that computational
efficiency is an intrinsic strength of the proposed formulation and
supports its use in repeated high-resolution forecasting, where the
upfront training cost can be amortized across many evaluations.

The computational formulation is also distinct from several recent
approaches to 3D microstructure learning. Graph-based methods represent
the evolving material through reduced descriptions of grains,
interfaces, or other structural entities
\cite{qin2024graingnn,fan2024gnn}, whereas other recent 3D frameworks
employ recurrent or latent-state temporal evolution
\cite{lanzoni2024extreme,razavi2026gcnlstm}. In contrast, the present
framework operates directly on dense volumetric fields and learns a
nonrecurrent finite-horizon mapping that produces multiple consecutive
full-field 3D states within each prediction block. These approaches
address different computational objectives and are therefore not
directly interchangeable. The particular advantage of the present
formulation lies in applications where complete volumetric fields and
their consecutive temporal evolution are required at high throughput.

Several limitations define the scope of the present study. The current
experiments consider a binary CH system with fixed
governing parameters, and therefore do not yet establish generalization
across material parameters, compositions, or different phase-field
models. In addition, all trajectories are generated numerically;
validation using experimentally measured 3D microstructures remains an
important extension. The reduced-context experiments employ leading
zero padding to maintain the fixed input length, which provides a
controlled test of limited temporal information but may introduce
transient artifacts. Finally, the CH residual is used as a
soft regularizer and therefore encourages, rather than exactly
enforces, the governing dynamics. These limitations motivate future
extensions toward parameter-conditioned modeling, variable-length
observation histories, experimentally informed prediction, and
uncertainty-aware data assimilation.

Overall, the results show that the combination of direct multi-frame
volumetric forecasting, factorized spatiotemporal representation, and
equation-based regularization provides an effective computational
strategy for 3D microstructure evolution. The framework is particularly
well suited to many-query settings in which complete 3D fields must be
predicted repeatedly, including accelerated simulation, parameter
estimation, optimization, and data-integrated materials modeling.

\section{Conclusion}
\label{sec:conc}
This work develops a physics-guided fully convolutional framework for
direct multi-frame prediction of 3D microstructure evolution. The
proposed formulation extends fully convolutional spatiotemporal
learning to native-grid volumetric phase-field data through shared 3D
spatial encoding and decoding and a factorized latent representation of
temporal, local spatial, and channel interactions. By learning a
finite-horizon evolution mapping, the framework predicts consecutive
future volumetric states directly rather than relying on sequential
hidden-state propagation. The numerical experiments demonstrate that the framework can
accurately represent 3D spinodal-decomposition dynamics under nominal
prediction, extended block-wise forecasting, and substantially reduced
temporal context. The results further show that physics guidance plays
an increasingly important role as the temporal information available to
the model decreases, providing additional dynamical structure that
improves robustness and preservation of interface-level morphology.
Because the physical residual is introduced only during training, these
benefits are obtained without changing the inference pathway. 
More broadly, this study establishes factorized fully convolutional
spatiotemporal learning as an effective high-throughput surrogate
strategy for dense 3D phase-field dynamics. The ability to predict
complete consecutive volumetric fields efficiently, while incorporating
governing-equation information into training, provides a foundation for
future extensions to parameterized and multicomponent systems,
experimentally measured 3D microstructures, uncertainty-aware
forecasting, and data-integrated materials modeling.

\section*{Declaration of competing interest} 
The authors declare that they have no known competing financial interests or personal relationships that could have appeared to influence the work reported in this paper.

\section*{Code and Data Availability}
The code developed for this work and the datasets generated and analyzed during the current study can be found here: \url{https://github.com/mtrimboli2018/CNN_STL_MicEvo}.

\section*{Declaration of generative AI and AI-assisted technologies in the manuscript preparation process} 
During the preparation of this work the author(s) used ChatGPT in order to improve language and readability. After using this tool/service, the author(s) reviewed and edited the content as needed and take(s) full responsibility for the content of the publication.

\section*{Acknowledgment}
X. Li's work was partially funded by the Division of Mathematical Sciences, National Science Foundation with the award numbers 2410678.

\bibliographystyle{elsarticle-num} 

\bibliography{reference}

\begin{thebibliography}{10}
\expandafter\ifx\csname url\endcsname\relax
  \def\url#1{\texttt{#1}}\fi
\expandafter\ifx\csname urlprefix\endcsname\relax\def\urlprefix{URL }\fi
\expandafter\ifx\csname href\endcsname\relax
  \def\href#1#2{#2} \def\path#1{#1}\fi

\bibitem{chen2002phasefield}
L.-Q. Chen, Phase-field models for microstructure evolution, Annual Review of
  Materials Research 32 (2002) 113--140.
\newblock \href {https://doi.org/10.1146/annurev.matsci.32.112001.132041}
  {\path{doi:10.1146/annurev.matsci.32.112001.132041}}.

\bibitem{rohrer2005interface}
G.~S. Rohrer, Influence of interface anisotropy on grain growth and coarsening,
  Annual Review of Materials Research 35~(1) (2005) 99--126.

\bibitem{deschamps2021precipitation}
A.~Deschamps, C.~R. Hutchinson, Precipitation kinetics in metallic alloys:
  Experiments and modeling, Acta Materialia 220 (2021) 117338.

\bibitem{voorhees1992ostwald}
P.~W. Voorhees, Ostwald ripening of two-phase mixtures, Annual Review of
  Materials Science 22 (1992) 197--215.
\newblock \href {https://doi.org/10.1146/annurev.ms.22.080192.001213}
  {\path{doi:10.1146/annurev.ms.22.080192.001213}}.

\bibitem{oono1988spinodal}
Y.~Oono, S.~Puri, Study of phase-separation dynamics by use of cell dynamical
  systems. i. modeling, Physical Review A 38~(1) (1988) 434--453.

\bibitem{boettinger2002solidification}
W.~J. Boettinger, J.~A. Warren, C.~Beckermann, A.~Karma, Phase-field simulation
  of solidification, Annual Review of Materials Research 32 (2002) 163--194.
\newblock \href {https://doi.org/10.1146/annurev.matsci.32.101901.155803}
  {\path{doi:10.1146/annurev.matsci.32.101901.155803}}.

\bibitem{tourret2022phasefield}
D.~Tourret, H.~Liu, J.~Llorca, Phase-field modeling of microstructure
  evolution: Recent applications, perspectives and challenges, Progress in
  Materials Science 123 (2022) 100810.
\newblock \href {https://doi.org/10.1016/j.pmatsci.2021.100810}
  {\path{doi:10.1016/j.pmatsci.2021.100810}}.

\bibitem{krill2002grain3d}
C.~E.~I. Krill, L.-Q. Chen, Computer simulation of 3-d grain growth using a
  phase-field model, Acta Materialia 50~(12) (2002) 3059--3075.
\newblock \href {https://doi.org/10.1016/S1359-6454(02)00084-8}
  {\path{doi:10.1016/S1359-6454(02)00084-8}}.

\bibitem{lifshitz1961kinetics}
I.~M. Lifshitz, V.~V. Slyozov, The kinetics of precipitation from
  supersaturated solid solutions, Journal of Physics and Chemistry of Solids
  19~(1--2) (1961) 35--50.
\newblock \href {https://doi.org/10.1016/0022-3697(61)90054-3}
  {\path{doi:10.1016/0022-3697(61)90054-3}}.

\bibitem{wagner1961theorie}
C.~Wagner, Theorie der alterung von niederschl{\"a}gen durch uml{\"o}sen
  (ostwald-reifung), Zeitschrift f{\"u}r Elektrochemie, Berichte der
  Bunsengesellschaft f{\"u}r physikalische Chemie 65~(7--8) (1961) 581--591.
\newblock \href {https://doi.org/10.1002/bbpc.19610650704}
  {\path{doi:10.1002/bbpc.19610650704}}.

\bibitem{akaiwa1994late}
N.~Akaiwa, P.~W. Voorhees, Late-stage phase separation: Dynamics, spatial
  correlations, and structure functions, Physical Review E 49 (1994)
  3860--3870.
\newblock \href {https://doi.org/10.1103/PhysRevE.49.3860}
  {\path{doi:10.1103/PhysRevE.49.3860}}.

\bibitem{wang2024systematic}
K.~G. Wang, X.~Li, Systematic and quantitative testing simulations and theories
  on phase coarsening by experiments, Materialia 37 (2024) 102192.
\newblock \href {https://doi.org/10.1016/j.mtla.2024.102192}
  {\path{doi:10.1016/j.mtla.2024.102192}}.

\bibitem{nestler2000multiphase}
B.~Nestler, A.~A. Wheeler, A multi-phase-field model of eutectic and peritectic
  alloys: numerical simulation of growth structures, Physica D: Nonlinear
  Phenomena 138~(1--2) (2000) 114--133.
\newblock \href {https://doi.org/10.1016/S0167-2789(99)00184-0}
  {\path{doi:10.1016/S0167-2789(99)00184-0}}.

\bibitem{stewart2020thinfilm}
J.~A. Stewart, R.~Dingreville, Microstructure morphology and concentration
  modulation of nanocomposite thin-films during simulated physical vapor
  deposition, Acta Materialia 188 (2020) 181--191.

\bibitem{cahn1958freeenergy}
J.~W. Cahn, J.~E. Hilliard, Free energy of a nonuniform system. i. interfacial
  free energy, The Journal of Chemical Physics 28~(2) (1958) 258--267.
\newblock \href {https://doi.org/10.1063/1.1744102}
  {\path{doi:10.1063/1.1744102}}.

\bibitem{cahn1961spinodal}
J.~W. Cahn, On spinodal decomposition, Acta Metallurgica 9~(9) (1961) 795--801.

\bibitem{shimokawabe2011dendrite}
T.~Shimokawabe, T.~Aoki, T.~Takaki, T.~Endo, A.~Yamanaka, N.~Maruyama,
  A.~Nukada, S.~Matsuoka, Peta-scale phase-field simulation for dendritic
  solidification on the {TSUBAME} 2.0 supercomputer, in: Proceedings of 2011
  International Conference for High Performance Computing, Networking, Storage
  and Analysis, 2011, pp. 1--11.

\bibitem{hunter2011phasefield3d}
A.~Hunter, F.~Saied, C.~Le, M.~Koslowski, Large-scale 3d phase field
  dislocation dynamics simulations on high-performance architectures, The
  International Journal of High Performance Computing Applications 25~(2)
  (2011) 223--235.

\bibitem{vondrous2014parallel}
A.~Vondrous, M.~Selzer, J.~H{\"o}tzer, B.~Nestler, Parallel computing for
  phase-field models, The International Journal of High Performance Computing
  Applications 28~(1) (2014) 61--72.

\bibitem{miyoshi2017ultralarge}
E.~Miyoshi, T.~Takaki, M.~Ohno, Y.~Shibuta, S.~Sakane, T.~Shimokawabe, T.~Aoki,
  Ultra-large-scale phase-field simulation study of ideal grain growth, npj
  Computational Materials 3 (2017) 25.
\newblock \href {https://doi.org/10.1038/s41524-017-0029-8}
  {\path{doi:10.1038/s41524-017-0029-8}}.

\bibitem{shi2017gpu}
X.~Shi, H.~Huang, G.~Cao, X.~Ma, Accelerating large-scale phase-field
  simulations with gpu, AIP Advances 7~(10) (2017).

\bibitem{du2020phasefield}
Q.~Du, X.~Feng, The phase field method for geometric moving interfaces and
  their numerical approximations, Handbook of Numerical Analysis 21 (2020)
  425--508.

\bibitem{zhang2020multiresolution}
X.~Zhang, K.~Garikipati, Machine learning materials physics: Multi-resolution
  neural networks learn the free energy and nonlinear elastic response of
  evolving microstructures, Computer Methods in Applied Mechanics and
  Engineering 372 (2020) 113362.

\bibitem{montesdeoca2021surrogate}
D.~Montes~de Oca~Zapiain, J.~A. Stewart, R.~Dingreville, Accelerating
  phase-field-based microstructure evolution predictions via surrogate models
  trained by machine learning methods, npj Computational Materials 7 (2021) 3.
\newblock \href {https://doi.org/10.1038/s41524-020-00471-8}
  {\path{doi:10.1038/s41524-020-00471-8}}.

\bibitem{yang2021selfsupervised}
K.~Yang, Y.~Cao, Y.~Zhang, S.~Fan, M.~Tang, D.~Aberg, B.~Sadigh, F.~Zhou,
  Self-supervised learning and prediction of microstructure evolution with
  convolutional recurrent neural networks, Patterns 2~(5) (2021) 100243.
\newblock \href {https://doi.org/10.1016/j.patter.2021.100243}
  {\path{doi:10.1016/j.patter.2021.100243}}.

\bibitem{farizhandi2023spatiotemporal}
A.~A.~K. Farizhandi, M.~Mamivand, Spatiotemporal prediction of microstructure
  evolution with predictive recurrent neural network, Computational Materials
  Science 223 (2023) 112110.
\newblock \href {https://doi.org/10.1016/j.commatsci.2023.112110}
  {\path{doi:10.1016/j.commatsci.2023.112110}}.

\bibitem{hu2022latent}
C.~Hu, S.~Martin, R.~Dingreville, Accelerating phase-field predictions via
  recurrent neural networks learning the microstructure evolution in latent
  space, Computer Methods in Applied Mechanics and Engineering 397 (2022)
  115128.
\newblock \href {https://doi.org/10.1016/j.cma.2022.115128}
  {\path{doi:10.1016/j.cma.2022.115128}}.

\bibitem{ahmad2023autoencoder}
O.~Ahmad, N.~Kumar, R.~Mukherjee, S.~Bhowmick, Accelerating microstructure
  modeling via machine learning: A method combining autoencoder and {ConvLSTM},
  Physical Review Materials 7~(8) (2023) 083802.

\bibitem{oommen2022learning}
V.~Oommen, K.~Shukla, S.~Goswami, R.~Dingreville, G.~E. Karniadakis, Learning
  two-phase microstructure evolution using neural operators and autoencoder
  architectures, npj Computational Materials 8 (2022) 190.
\newblock \href {https://doi.org/10.1038/s41524-022-00876-7}
  {\path{doi:10.1038/s41524-022-00876-7}}.

\bibitem{li2020fno}
Z.~Li, N.~Kovachki, K.~Azizzadenesheli, B.~Liu, K.~Bhattacharya, A.~Stuart,
  A.~Anandkumar, Fourier neural operator for parametric partial differential
  equations, in: International Conference on Learning Representations, 2021.

\bibitem{wang2018predrnn}
Y.~Wang, Z.~Gao, M.~Long, J.~Wang, P.~S. Yu, Predrnn++: Towards a resolution of
  the deep-in-time dilemma in spatiotemporal predictive learning, in:
  Proceedings of the 35th International Conference on Machine Learning, 2018,
  pp. 5123--5132.

\bibitem{wang2018e3dlstm}
Y.~Wang, L.~Jiang, M.-H. Yang, L.-J. Li, M.~Long, L.~Fei-Fei, Eidetic 3d lstm:
  A model for video prediction and beyond, in: International Conference on
  Learning Representations, 2019.

\bibitem{raissi2019pinn}
M.~Raissi, P.~Perdikaris, G.~E. Karniadakis, Physics-informed neural networks:
  A deep learning framework for solving forward and inverse problems involving
  nonlinear partial differential equations, Journal of Computational Physics
  378 (2019) 686--707.

\bibitem{trimboli2026fully}
M.~Trimboli, M.~Alsubaie, S.~M. Perera, K.-G. Wang, X.~Li, Fully convolutional
  spatiotemporal learning for microstructure evolution prediction, Journal of
  Materials Science 61 (2026) 24967--24991.
\newblock \href {https://doi.org/10.1007/s10853-026-13194-w}
  {\path{doi:10.1007/s10853-026-13194-w}}.

\bibitem{trimboli2026physicsguided}
M.~Trimboli, W.~Liu, X.~Li, Physics-guided fully convolutional spatiotemporal
  learning toward digital-twin-enabled microstructure evolution prediction,
  arXiv preprint arXiv:2606.20983 (2026).
\newblock \href {http://arxiv.org/abs/2606.20983} {\path{arXiv:2606.20983}}.

\bibitem{alkemper2001dendritic}
J.~Alkemper, P.~W. Voorhees, Three-dimensional characterization of dendritic
  microstructures, Acta Materialia 49~(5) (2001) 897--902.
\newblock \href {https://doi.org/10.1016/S1359-6454(00)00355-4}
  {\path{doi:10.1016/S1359-6454(00)00355-4}}.

\bibitem{zhang2024fecr3d}
T.~Zhang, J.~Zhong, L.~Zhang, Multi-objective optimization assisting
  three-dimensional quantitative cahn--hilliard simulations of microstructure
  evolution in fe--cr alloys during spinodal decomposition, Computational
  Materials Science 244 (2024) 113260.
\newblock \href {https://doi.org/10.1016/j.commatsci.2024.113260}
  {\path{doi:10.1016/j.commatsci.2024.113260}}.

\bibitem{qin2024graingnn}
Y.~Qin, S.~DeWitt, B.~Radhakrishnan, G.~Biros, Graingnn: A dynamic graph neural
  network for predicting 3d grain microstructure, Journal of Computational
  Physics 510 (2024) 113061.
\newblock \href {https://doi.org/10.1016/j.jcp.2024.113061}
  {\path{doi:10.1016/j.jcp.2024.113061}}.

\bibitem{fan2024gnn}
S.~Fan, A.~L. Hitt, M.~Tang, B.~Sadigh, F.~Zhou, Accelerate microstructure
  evolution simulation using graph neural networks with adaptive spatiotemporal
  resolution, Machine Learning: Science and Technology 5~(2) (2024) 025027.
\newblock \href {https://doi.org/10.1088/2632-2153/ad3e4b}
  {\path{doi:10.1088/2632-2153/ad3e4b}}.

\bibitem{lanzoni2024extreme}
D.~Lanzoni, A.~Fantasia, R.~Bergamaschini, O.~Pierre-Louis, F.~Montalenti,
  Extreme time extrapolation capabilities and thermodynamic consistency of
  physics-inspired neural networks for the 3d microstructure evolution of
  materials via cahn--hilliard flow, Machine Learning: Science and Technology
  5~(4) (2024) 045017.
\newblock \href {https://doi.org/10.1088/2632-2153/ad8532}
  {\path{doi:10.1088/2632-2153/ad8532}}.

\bibitem{razavi2026gcnlstm}
H.~Razavi, N.~Moelans, Physics-informed gcn-lstm framework for long-term
  forecasting of 2d and 3d microstructure evolution, npj Computational
  Materials 12 (2026) 190.
\newblock \href {https://doi.org/10.1038/s41524-026-01999-x}
  {\path{doi:10.1038/s41524-026-01999-x}}.

\bibitem{gao2022simvp}
Z.~Gao, C.~Tan, L.~Wu, S.~Z. Li, Simvp: Simpler yet better video prediction,
  in: Proceedings of the IEEE/CVF Conference on Computer Vision and Pattern
  Recognition, 2022, pp. 3170--3180.
\newblock \href {https://doi.org/10.1109/CVPR52688.2022.00317}
  {\path{doi:10.1109/CVPR52688.2022.00317}}.

\bibitem{tan2025simvpv2}
C.~Tan, Z.~Gao, S.~Li, S.~Z. Li, Simvpv2: Towards simple yet powerful
  spatiotemporal predictive learning, IEEE Transactions on Multimedia 27 (2025)
  5170--5184.
\newblock \href {https://doi.org/10.1109/TMM.2025.3543051}
  {\path{doi:10.1109/TMM.2025.3543051}}.

\bibitem{tolstikhin2021mlp}
I.~Tolstikhin, N.~Houlsby, A.~Kolesnikov, L.~Beyer, X.~Zhai, T.~Unterthiner,
  J.~Yung, A.~Steiner, D.~Keysers, J.~Uszkoreit, et~al., Mlp-mixer: An all-mlp
  architecture for vision, arXiv preprint arXiv:2105.01601 1~(2) (2021) 3.

\bibitem{soares2023exponentialintegratorsphasefieldequations}
E.~do~A.~Soares, A.~G.~B. Jr., F.~W. Tavares,
  \href{https://arxiv.org/abs/2305.08998}{Exponential integrators for
  phase-field equations using pseudo-spectral methods: A python implementation}
  (2023).
\newblock \href {http://arxiv.org/abs/2305.08998} {\path{arXiv:2305.08998}}.
\newline\urlprefix\url{https://arxiv.org/abs/2305.08998}

\bibitem{ZWang04}
Z.~Wang, A.~Bovik, H.~Sheikh, , E.~P. Simoncelli, Image quality assessment:
  from error visibility to structural similarity, IEEE Trans. Image Process 13
  (2004) 600--612.

\bibitem{gibbs2002onchoosing}
A.~L. Gibbs, F.~E. Su, On choosing and bounding probability metrics,
  International Statistical Review 70~(3) (2002) 419--435.

\end{thebibliography}

\end{document}